\documentclass[10pt,twocolumn,letterpaper]{article}

\usepackage[pagenumbers]{cvpr} % To force page numbers, e.g. for an arXiv version

\usepackage[dvipsnames]{xcolor}

\usepackage{graphicx}
\usepackage{comment}
\usepackage{amsmath}
\usepackage{amssymb}
\usepackage{amsfonts}
\usepackage{pifont}
\usepackage{booktabs}
\usepackage{algorithmic}
\usepackage{textcomp}
\usepackage{cite}
\usepackage{xcolor,soul}
\usepackage[normalem]{ulem}
\usepackage{kotex}
\usepackage{multirow, array}
\usepackage{url}
\usepackage{footnote}
\makesavenoteenv{tabular}
\usepackage{relsize}
\usepackage[linesnumberedhidden,ruled,vlined]{algorithm2e}
\usepackage{adjustbox}
\usepackage{caption}
\usepackage{stfloats}
\usepackage[accsupp]{axessibility}
\usepackage[symbol]{footmisc}

\definecolor{cvprblue}{rgb}{0.21,0.49,0.74}
\usepackage[pagebackref,breaklinks,colorlinks,citecolor=cvprblue]{hyperref}
\newcommand{\cmark}{\ding{51}}%
\newcommand{\xmark}{\ding{55}}%
\def\paperID{*****} % *** Enter the Paper ID here
\def\confName{CVPR}
\def\confYear{2024}

\title{UOD: Unseen Object Detection in 3D Point Cloud}

\author{Hyunjun Choi$^1,^{2\textsuperscript{*}}$ \quad
Daeho Um$^1$ \quad
Hawook Jeong$^2$ \\
{\small$^1$ ASRI, ECE., Seoul National University} \quad
{\small$^2$ RideFlux Inc.} \\
{\tt\footnotesize numb7315@snu.ac.kr} \quad
{\tt\footnotesize daehoum1@snu.ac.kr
} \quad
{\tt\footnotesize hawook@rideflux.com}
}

\begin{document}
\maketitle

Existing 3D object detectors encounter extreme challenges in localizing unseen 3D objects and recognizing them as unseen, which is a crucial technology in autonomous driving in the wild. 
To address these challenges, we propose practical methods to enhance the performance of 3D detection and Out-Of-Distribution (OOD) classification for unseen objects. 
The proposed methods include anomaly sample augmentation, learning of universal objectness, learning of detecting unseen objects, and learning of distinguishing unseen objects. To demonstrate the effectiveness of our approach, we propose the KITTI Misc benchmark and two additional synthetic OOD benchmarks: the Nuscenes OOD benchmark and the SUN-RGBD OOD benchmark.
The proposed methods consistently enhance performance by a large margin across all existing methods, giving insight for future work on unseen 3D object detection in the wild.
%Open world에 대해서 신뢰할 수 있는 물체 탐지기를 구축하여 OOD(Out-of-Distribution) 객체를 감지하는 것은 중요하지만 미개척된 분야입니다. 특히, pointcloud 상에서 3D object detection 에 대해
%In this paper, we raise a new issue on Unidentified Foreground Object (UFO) detection in 3D point clouds, which is a crucial technology in autonomous driving in the wild. UFO detection is challenging in that existing 3D object detectors encounter extremely hard challenges in both 3D localization and Out-of-Distribution (OOD) detection. 
% To tackle these challenges, we suggest a new UFO detection framework including three tasks: evaluation protocol, methodology, and benchmark.
% The evaluation includes a new approach to measure the performance on our goal, i.e. both localization and OOD detection of UFOs. The methodology includes practical techniques to enhance the performance of our goal. The benchmark is composed of the KITTI Misc benchmark, Nuscenes OOD benchmark, and our synthetic benchmark for modeling a more diverse range of UFOs. 
% The proposed framework consistently enhances performance by a large margin across all four baseline detectors: SECOND, PointPillars, PV-RCNN, and PartA2, giving insight for future work on UFO detection in the wild.

\section{Introduction}
% 기존연구 방향

% 자율주행 상황에서 lidar에 대한 3D Object detection은 핵심적인 인식 기술이다.
% 3D object detector 의 인식 성능은 고도화를 이루었지만, 이들을 실제 application 적용에 있어 안정성에 대한 확신은 여전히 부족하다. 특히, 3D object detector가 Unidentified foreground object 또는 unknown object에 대해 assigning high confidence scores 하는 문제가 하나의 예시이다. 
% 최근, Image 상에서 2D object detection 대해 OOD detection 이나 open set object detection 등이 이러한 문제를 다루고 있다. 비슷하게, 최근 point cloud 상 3D object detection에서도 이를 비슷하게 다루려고 하고있다.
In autonomous driving scenarios, 3D object detection using point clouds is a crucial perception technology. 
The recognition performance of 3D object detectors has advanced under the closed-world assumption, but there are two issues regarding stability in open-world scenarios: whether they can detect unseen objects and differentiate them as unseen objects.
In first issue, methods addressing open-world 2D object detection~\cite{joseph2021towards,kim2022learning} or instance segmentation~\cite{saito2022learning} 
have tackled issues in 2D images. However, there are still uncertainties regarding 3D point clouds.
In the second issue, methods addressing Out-of-Distribution (OOD) detection~\cite{du2022unknown, du2022vos} for 2D object detection on images have tackled similar challenges. Similarly, in the realm of 3D object detection~\cite{huang2022out,wong2020identifying,cen2021open} on point clouds, efforts are underway to address these issues, but they only focus on the classification of unseen objects. Detection of unseen objects is a prerequisite preceding classification for safety in autonomous driving. In this paper, the term `OOD classification' will replace the conventional term `OOD detection'.
% 문제점 및 원인 : localization 어려움, 3D OD의 특수성
% 하지만,  우리는 3D object detector가 Unidentified foreground object 에 대한  OOD detection이 뿐만 아니라, 특히 localization 에 있어 많은 어려움이 있다는 점을 발견하였다. 2D image에서와 달리 Lidar point cloud 는 sparse 하여 다양한 size를 갖는 Unidentified foreground object에 대해 정확한 context를 얻고 이를 정확하게 localization 을 하는 것은 challenging 하다. 
% intro figure

% \begin{figure}[t!]
%     \centering
%     % \includegraphics[width=1.0\linewidth]{figure/intro_fig.pdf}
%     % \includegraphics[width=1.0\linewidth]{figure/intro_fig2.pdf}
%     \includegraphics[width=1.0\linewidth]{figure/intro_fig3.pdf}
    
%     \vspace{-0.1cm}
%     \caption{\textbf{Enhancement of trade-off by the proposed method.} The proposed three factors  improve both OOD performance (AUROC) and accuracy
%      (ACC) when they are added to the 
%     %than 
%     baseline OE.}
%     \label{intro_fig}
% \end{figure}

%%%%%%%%%
% sub figure
\begin{figure}[t!]
%\captionsetup{font=footnotesize}
\centering
\begin{subfigure}[b]{0.85\linewidth}
        \caption{}
        \includegraphics[width=\linewidth]{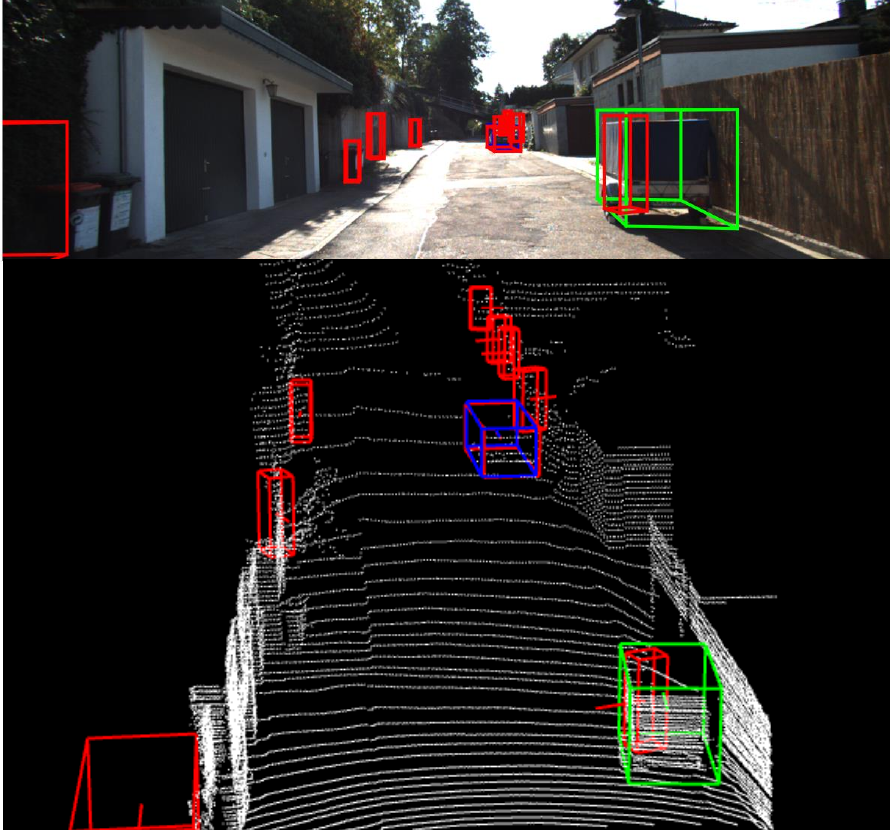}
        \label{intro_a}
\end{subfigure}
\begin{subfigure}[b]{0.85\linewidth}
        \caption{}
        \includegraphics[width=\linewidth]{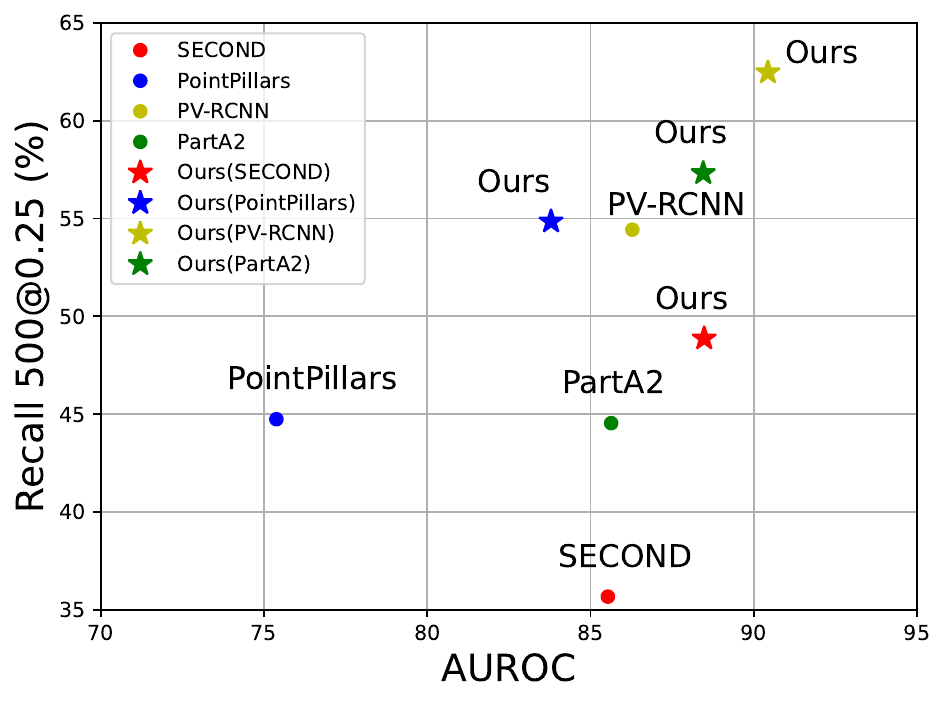}
        \label{intro_b}
\end{subfigure}

\caption{ \textbf{Base 3D object detector and our method comparison.} (a): 3D object detection result of baseline SECOND~\cite{yan2018second} on KITTI~\cite{geiger2012we} `Misc' class object; (b): Comparison of the base detector and our method in two aspects: unseen object detection performance (Recall) and OOD classification performance (AUROC).
}
\label{intro_fig}
\end{figure}

We have found that recent 3D object detectors~\cite{yan2018second, lang2019pointpillars, shi2019part, shi2020pv} face challenges in detecting unseen objects in open-world scenarios. Specifically, unlike 2D images, for the 3D detection of unseen objects, % Unlike 2D images, 
Lidar point clouds are sparse, making it challenging to obtain accurate context and precisely localize unseen objects with various sizes. As depicted in Figure~\ref{intro_a}, a SECOND~\cite{yan2018second} detector, which is trained on the classes of Car, Pedestrian, and Cyclist, fails to localize the `Misc' class object within the green box even at a close distance. Instead, SECOND recognizes the unseen object as a smaller pedestrian, leading to significant failures in 3D localization.
Furthermore, addressing localization issues is a crucial prerequisite for OOD classification. If the detector fails to localize an unseen object, obtaining its detection results and classifying it as an unseen object becomes impossible. Therefore, in open-world scenarios, it is essential to evaluate not only the OOD classification but also their detection performance.

In this paper, we address the open-world 3D object detection problem through two main directions: (i) introducing an integrated protocol for evaluating the safety of 3D object detectors, and (ii) presenting methodologies for enhancement.
We propose a comprehensive protocol for evaluating open-world 3D object detection, assessing both the detection of unseen objects and their OOD classification, simultaneously. Our ideal 3D object detector excels in precisely localizing unseen objects while assigning high OOD confidence scores to them. As in previous open-world object detection~\cite{kim2022learning}, the detection performance is measured by the recall of unseen objects at multiple IOU thresholds. OOD classification utilizes conventional base metrics: AUROC, FPR, and AUPR.

In line with our integrated 3D detector evaluation protocol, we propose methods to simultaneously enhance unseen object detection and OOD classification performance. We introduce an anomaly sample augmentation approach inspired by the outlier exposure method~\cite{hendrycks2018deep}, acquiring anomaly samples from indoor scene SUN-RGBD~\cite{song2015sun} data and incorporating them as a new additional class for training. As a result, our method undergoes training to detect unseen objects of various sizes.
Next, we address the conflicting aspects between unseen object detection and OOD classification. In the structure of existing 3D detectors, assigning low In-Distribution (ID) confidence scores to unseen objects is inherently linked to assigning low objectness scores for unseen objects. We need to assign low In-Distribution (ID) confidence scores to unseen objects while assigning high objectness scores for detection.
Therefore, we add a separate objectness node alongside the classification nodes for the 3D object detector.
In addition to the proposed augmentation, we introduce a novel technique to enhance OOD classification performance by leveraging energy-based regularization and outlier-aware supervised contrastive learning using the anomaly samples introduced in the proposed augmentation.
As evident from Figure~\ref{intro_b}, the application of our methods yields improvements in both the detection and OOD classification of unseen objects compared to the four baseline detectors.
% 해결책 3: Unidentified Foreground Object 데이터의 부족 및 benchmark 부족 
%  새로운 benchmark 제시

% 마지막으로, 우리는 기존 KITTI 데이터에서 UFO 로써 'Misc' class object를 선정한 KITTI Misc benchmark에 대해 evaluation을 실시한다. 하지만, 해당 object outdoor scene에서 주로 나타나는 UFO에 해당한다.
% 우리는 in-door scene 에 있는 다양한 새로운 물체를 out-door scene KITTI에 출연시킴으로써, 더 다양한 UFO 에 대해 safety를 평가하는 benchmark를 제안한다. 제안한 synthetic benchmark 는 Sun-RGBD 데이터 중에 augmentation에 활용하지 않은 class를 활용하여 구성한다. 더나아가, 우리는 challenging 한 benchmark 구성을 위해 in-door scene 의 object를 바로 out-door scene에 출연시키는 것이 아니라, out-door scene과의 domain gap 을 줄이는 Nearest Neighbor grid sampling 방법을 거쳐 출연시킨다. 그 결과, 우리는 기존 baseline detector에 대해서 OOD detection 관점에서 더 challenging한 benchmark를 구성할수있다.

Practically, we measure unseen object detection and OOD classification on Lidar-based detectors trained on KITTI~\cite{geiger2012we} scenes.
We designate the `Misc' class as the unseen object, creating the KITTI Misc benchmark, and propose baselines for four existing detectors: SECOND~\cite{yan2018second}, PointPillars~\cite{lang2019pointpillars}, PV-RCNN~\cite{shi2020pv}, and PartA2~\cite{shi2019part}.
Moreover, to assess safety for a more diverse range of unseen objects, we propose two additional benchmarks: the Nuscenes~\cite{caesar2020nuscenes} OOD benchmark and the SUN-RGBD~\cite{song2015sun} OOD benchmark. We introduce various new unseen objects from large-scale outdoor scenes Nuscenes and indoor scenes SUN-RGBD into the outdoor scene of KITTI.  
% The Nuscenes OOD benchmark is constructed by introducing point clouds of five OOD object classes in Nuscenes that are not observed in KITTI scenes into the KITTI scenes themselves. Nuscenes OOD benchmark is a realistic benchmark that models a wider variety of OOD objects in outdoor scenes.
% The synthetic benchmark utilizes indoor scene objects which are classes from the Sun-RGBD dataset that have not been used for augmentation.
% Furthermore, for the construction of a challenging benchmark, we employ the Nearest Neighbor grid Sampling method to reduce the domain gap between indoor and outdoor scenes, ensuring that in-door scene objects are incorporated into the outdoor scene. As a result, we can create a more challenging benchmark for OOD detection from the perspective of existing baseline detectors.

In summary, our contribution can be outlined as follows:
(i) Introducing an integrated protocol for evaluating open-world 3D object detection on KITTI scenes, providing baseline assessments for four 3D object detectors: SECOND, PointPillars, PV-RCNN, and PartA2.
(ii) Applying practical methods enhances open-world 3D object detection performance in both unseen object detection and OOD classification from existing 3D object detector baselines.
(iii) Constructing three benchmark scenarios to model a diverse range of unseen objects and confirming our method's significant improvement over baselines
        \section{Related Work}

\noindent\textbf{Unseen Object Detection in Open-world}  

Unseen object detection has primarily been tackled within the realm of 2D object detection under the open-world assumption. ~\cite{kim2022learning} advanced universal unseen object detection by omitting classification nodes within the region proposal network of 2D object detectors. ~\cite{joseph2021towards} refines unseen object detection within open-world contexts through incremental learning of unseen objects. ~\cite{saito2022learning} endeavors to enhance unseen object detection by substituting unannotated objects with annotated objects via augmentation in the instance segmentation task. While existing works have addressed unseen object detection for 2D images, we have discovered significant challenges in 3D object detection on 3D point clouds and have made substantial improvements in this regard.
% Open-Set Object Detection (OSOD) extends from object detection to Open Set Recognition (OSR)~\cite{scheirer2012toward}.
% OSOD is formally introduced in~\cite{dhamija2020overlooked}, evaluating detectors like Faster-RCNN~\cite{ren2015faster}, Retinaet~\cite{lin2017focal}, and YOLO~\cite{redmon2016you}. Their key protocol, wilderness, measures the precision ratio between scenes with mixed unknown and purely known instances.
% Recently, OpenDet~\cite{han2022expanding} proposes to expand low-density latent regions to improve OSOD.
% However, these approaches require scenes with mixed unknown and purely known. 
% Similarly, OSOD in 3D object detection~\cite{cen2021open} aimed to evaluate OSOD when known and unknown instances coexist. However, they use heuristic confidence score thresholds for unknown instances.
% Our protocol is more practical since it can be applied to individual scenes with mixed unknown instances. It evaluates two aspects of the unknown elements present in individual scenes: localization and OOD detection.

\noindent\textbf{OOD detection on Object Detection}

% OOD detection~\cite{hendrycks2016baseline} is similar to the rejection of unknown classes in OSR~\cite{scheirer2012toward} but does not require keeping the accuracy of known classes. 따라서, 조금 더 간단한 protocol 을 가진다. Open set recogntion 에서는 OOD 와 함께 있는 scene과 없는 scene이 필요하다. 하지만, 우리는 둘이 공존하는 practical한 setting에서 OOD detection을 수행하는 점이 다르다.
% 최근 2D object detection 에서 STUD~\cite{du2022unknown} 에서 처음으로 이러한 OOD detection 에 대한 VOS~\cite{du2022vos} 논문에서 이러한 protocol 이 제시되었따.
% 단순하게, object가 없는 scene에 대해서 얻은 score간의 AUROC를 측정한다.
% 하지만, 이는 practical한 setting 에 적합하지 않다.
% Lidar 3D point cloud 상에서 이러한 OOD detection을 다루는 논문ㅇ~\cite{huang2022out}
In recent 2D object detection, the STUD~\cite{du2022unknown} and VOS~\cite{du2022vos} papers introduced OOD classification tasks. They measure the OOD classification performance by distinguishing scenes with only known (seen) objects and scenes without them, considering all scores obtained from the detector. However, these methods deal only with 2D images and may not be practical as they are not suitable for many real-world environments where known (seen) and unknown (unseen) objects coexist. 
%우리의 방법은 known과 %unknown이 coexist할때 적용가능한 실용적인 protocol 이다.
% 최근, Lidar 3D point cloud 상에서  known과 unknown이 coexist 하는경우에 OOD detection을 평가하고자 한 논문~\cite{huang2022out}이 있었다. 하지만, 이는 unknown instance에 대해 iou threshold 를 heuristic하게 주어 
% 이에 대한 OOD score를 얻는 방식을 쓴다. 해당 방법은 하나의 detector에 대해 효과적이지만 서로 localizatoin을 다르게하는 여러 detector들에 대해 일괄적인 OOD detection성능을 얻음에 어려움이있다. 우리의 방법은 heuristic없이 one-to-one matching을 기반으로 여러 deteector에 대해 일관성있는 OOD detection 성능을 내고자한점이 다르다. 
Recently, in Lidar-based 3D object detection, an OOD classification task has been proposed in~\cite{huang2022out} that aimed to evaluate OOD classification when known (seen) and unknown (unseen) instances coexist. However, this method solely focuses on OOD object dataset generation without improving performance. Similarly, 3D Open-set object detection~\cite{cen2021open} aimed to improve detection accuracy when known and unknown instances coexist. However, due to the different metrics, direct comparisons become difficult, and unlike using heuristics to distinguish unknown instances, we provide a heuristic-free clear evaluation.
% \noindent\textbf{Localization cues for object detection}
% Learning to Detect Every Thing
% in an Open World~\cite{kim2022learning}
% Faster-RCNN의 region proposal network에 대해 unknown instance에 대해 proposal 을 형성함에 있어
% localization을 측정하고자한다.
% Learning to detect everything in an open world~\cite{saito2022learning}
% Mask-RCNN 기반의 instance segmentation 에 대해  unknown instance 에 대해 이를 수행하도록하는 데 목표를 두고있다.
% 우리의 방법은 이들에서 강조하는 open-world에 대해 localization 하고자하는 점에서 공통점이 있으나,
% 우리는 context가 없거나 sparse 하여 challenging한  point cloud 에 대해서 3D unknown instance를 localization 한 다는점에서 다르다. 3D object detection 에 대해 이러한 시도는 아는바로 최초이다.

\noindent\textbf{Lidar-based 3D Object Detection}

3D object detection based on Lidar point clouds has seen significant improvement by aggregating features through voxel-based learning~\cite{zhou2018voxelnet}.
SECOND~\cite{yan2018second} enhances speed over VoxelNet~\cite{zhou2018voxelnet} by replacing its conventional 3D convolution with sparse convolution. PointPillars~\cite{lang2019pointpillars} divides the point cloud into pillar units and applies PointNet to each unit. In contrast to SECOND and VoxelNet, which use 3D convolution to integrate voxel units, PointPillars uses 2D convolution to integrate pillar units which boosts efficiency in time. PartA2~\cite{shi2019part} newly designs a RoI-aware point cloud pooling module to encode effective features of 3D proposals. PV-RCNN~\cite{shi2020pv} extends SECOND, preserving more 3D structure information by adding a keypoint branch. 
% 기존 방법들은 오로지 in-distribution data에 대해 detection precision 을 하는데 집중을 해왔다
% 하지만, 이들의 OOD data 또는 unidentified foreground object 에 대해 어느정도 구분을 해내며 localization 을 해내는지에 대해서 아직 명확히 조사된적이 없었다.
Existing methods have primarily focused on improving detection precision in closed environments. However, there has not been a clear investigation into the ability of 3D object detectors in open-world scenarios to detect and differentiate unseen objects.

\section{3D Open-World Object Detection}

%OUtline of the section. ...........
\subsection{Problem Formulation and Evaluation}
\begin{figure}[t!]
    \centering
    \includegraphics[width=0.9\linewidth]{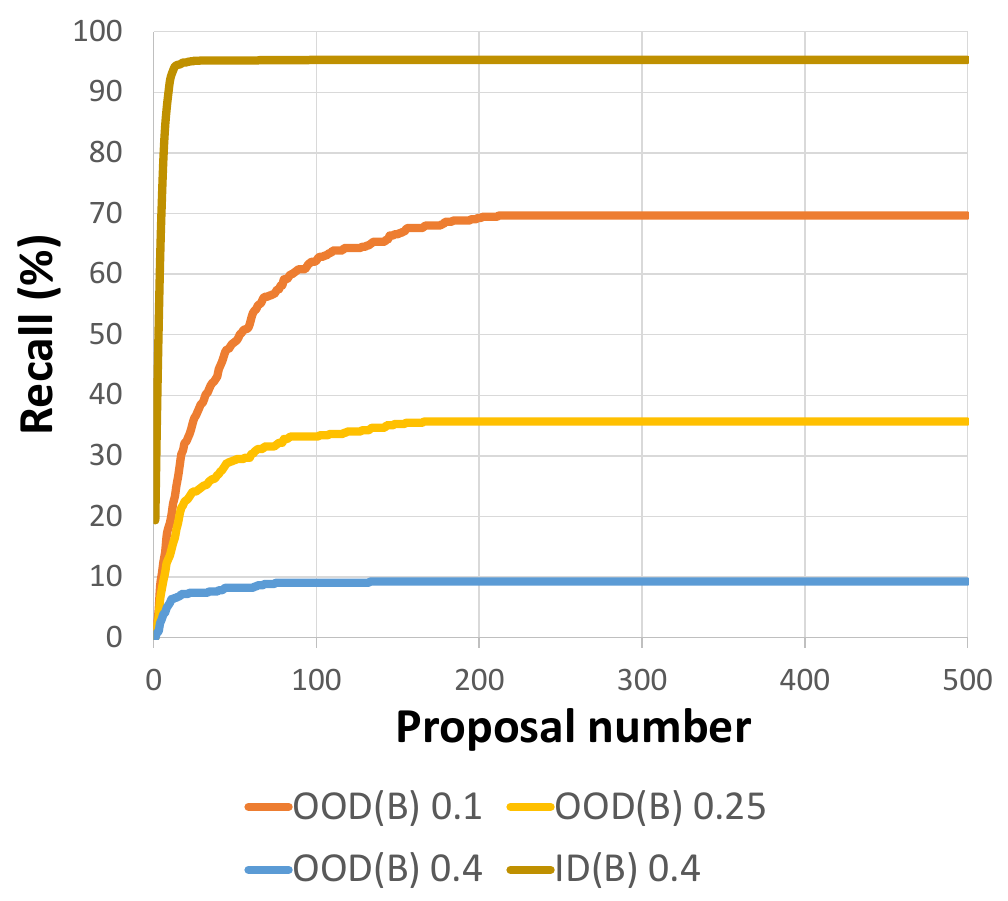}
    \caption{ \textbf{In-Distribution (ID)(seen) and Out-Of-Distribution (OOD)(unseen) object localization performance comparison.} This plot illustrates the recall for both ID (seen) and OOD (unseen) objects based on the proposal number. This depicts the recall for OOD objects at IoU thresholds of 0.1, 0.25, and 0.4.}
    \label{ID_OOD_localization}
 %   \vspace{-0.4cm}
\end{figure}
 We can formalize a Lidar-based 3D object detector that maps a point cloud consisting of $M$ points to $N$ detected objects as $\mathbf{z}(\mathbf{x}):\mathbb{R}^{D\times M}\to\mathbb{R}^{L\times N}$
    , where $D$ denotes the dimension of a point including location $(x,y,z)$ and $L$ denotes the dimension of detection results. 
    Object detection results consist of $N$ detected objects $\{O_{1},\ldots,O_{N}\}$ where $O_{i}=\{conf_{i}, \mathbf{c}_{i}, x_{i},y_{i},z_{i}, l_{i}, w_{i}, h_{i}, \theta_{i} \}$. Here, $conf_{i}$ is defined as the final objectness score, representing the degree to which an object is present.
A classification score with a total of $K$ classes is defined as $\mathbf{c}_{i}=[c_{1}, c_{2}, \ldots, c_{K}]^\top$.
The set $\{x_{i}, y_{i}, z_{i}, l_{i}, w_{i}, h_{i}, \theta_{i}\}$ corresponds to the 3D detection box, defined as a cuboid with an orientation angle.

In practical terms, to address the unseen object detection problem, we utilize a detector that identifies three classes in the KITTI dataset~\cite{geiger2012we}, including car, pedestrian, and cyclist (\textit{i.e.}, $K=3$). For unseen objects, we define the `Misc' class provided in the actual KITTI dataset. We refer to this as the KITTI Misc benchmark and propose a protocol for its evaluation.
In the evaluation, we simultaneously assess two aspects of unseen objects: unseen object detection and OOD classification. For localization, we utilize $conf_{i}$. For OOD classification, we obtain scalar scores (e.g., MSP~\cite{hendrycks2016baseline}, Energy~\cite{liu2020energy}) from $\mathbf{c}_{i}$ for evaluation. Unless stated otherwise, we use Energy score~\cite{liu2020energy} for evaluation in this paper.

\subsubsection{Evaluation of Unseen Objects Detection}
% 일반적으로, recall은 object detector의 safety를 위해 필수적인 metric이다.
% 실제 KITTI 에서의 detector들은 주로 최대 500개의 결과를 얻는 base setting을 따른다. 
% 우리는 실제 KITTI 에서의 SECOND detector에 대해 recall 결과를 도시한다.
% 구체적으로, recall은 proposal number와 iou threshold 기준으로 측정을 한다. 예측결과를 일괄적으로 proposal $conf_{i}$  를 기준으로  proposal number $T$에 따라 Top-$T$ 의 개수로 제한을 하고 마찬가지로 IOU threshold를 기준으로 찾은것으로 하여 실제 object에 대해 이를  예측으로 찾아낸 TP를 계산하여 $Recall=\frac{TP}{TP+FN}$을 계산한다. 실직적으로, 우리는 detection결과는 최종적으로 최대 $N=500$으로 통일하여 500개 까지의 proposal number 에 따른 recall을 측정하였다.
% 그래프에서 보듯이, 기존 baseline detector SECOND는 OOD의 localization에 있어서 같은 threshold 0.40에서 IND 에 비해 크게 떨어짐을 확인한다. 
% 그리고, 그래프에서 보듯이 우리의 recall은 proposal number 에 대해서 300 이후에  큰차이가 나지 않으므로, proposal number는 500으로 고정하고  IOU threshold로는 0.10, 0.25, 0.40 세가지에 대해 측정함으로써 localization 성능을 평가한다.

Generally, recall is a crucial metric for ensuring the safety of an object detector. As in previous open-world object detection studies~\cite{kim2022learning}, localization performance on OOD objects can be measured by recall.
In actual KITTI settings, detectors often follow a base setting, obtaining a maximum of 500 results. We demonstrate recall results for the actual SECOND detector on KITTI as described in Fig~\ref{ID_OOD_localization}. Specifically, recall is measured based on the proposal number and IOU threshold criteria. The predictions are uniformly restricted to the top-$k$ based on the score $conf_{i}$ and similarly found based on the IOU threshold, calculating True Positives (TP) for objects predicted among actual objects, and then computing $Recall=\frac{TP}{TP+FN}$. As evident from the graph, the baseline detector, SECOND, significantly lags behind in OOD localization compared to ID at the same threshold of 0.40. Furthermore, our recall in the graph shows minimal differences beyond a proposal number of 300. Therefore, we fix the proposal number $k=500$ and evaluate localization performance based on three IOU thresholds: 0.10, 0.25, and 0.40.
\subsubsection{Evaluation of OOD Classification}
% % 우리는 최종적인 detection result 기반으로 
% % ID 에 대한 classification score와 OOD 에 대한 classification score에 대한 scalar를 얻어 OOD detection을 수행한다~\cite{hendrycks2016baseline}. AUROC, FPR95, AUPR이다.
% % 기존 논문~\cite{huang2022out} 에서는 OOD data에 대해 iou threshold 0.3이상인것을 얻어 score를 얻고자 하지만, 기존의 문제는 동일한 하나의 detector에 대해 평가함에 문제없지만 여러가지의 detector에 대해 일괄적으로 적용시에 OOD data의 수가 달라짐으로써 어려움이 있다.
% % 따라서, 우리는 detector에 관계없이 통합적으로 OOD를 측정하기 위해 gt 에 대해 detection result를 one-to-one matching을 하는 알고리즘을 제안한다. 우리의 알고리즘은 DETR~\cite{carion2020end} 에서와 비슷하게 bipartite matching을 최적화하는 hungarian algorithm을 기반으로 한다.
% %\input{table_latex/B_algorithm}
We obtain scores of object instances from the detection results where there is overlap, following the approach of OOD classification in the existing methods for both 2D object detectors~\cite{du2022vos,du2022unknown} and 3D object detectors~\cite{huang2022out}. Detailed algorithms are covered in the supplementary material. Subsequently, we adhere to the baseline method~\cite{hendrycks2016baseline} for OOD classification in conventional image classification tasks. The evaluation metrics include AUROC, FPR95, and AUPR.
\subsection{Proposed Method}

Baseline 3D object detectors struggle with the detection and OOD classification of unseen objects. To address this, we employ two key strategies. First, inspired by outlier exposure~\cite{hendrycks2018deep}, we introduce auxiliary unseen object data by copying and pasting from SUN-RGBD~\cite{song2015sun} indoor scenes, treating it as a new `Anomaly' class for training unseen object detection across various sizes. Figure~\ref{fig3_a} illustrates this sample from SUN-RGBD.
Second, to boost the OOD classification performance of baseline detectors, we utilize the Anomaly-class data to apply energy-based regularization and outlier-aware contrastive learning. Our approach consists of four main techniques: (i) Anomaly Sample Augmentation, (ii) Learning on Objectness, (iii) Learning on Localizing Unseen Objects, and (iv) Learning on Distinguishing Unseen Objects.

\subsubsection{Anomaly Sample Augmentation}
In the existing SECOND, the augmentation method during training involves sampling ground truths from the database, specifically copying object points and labels from the ground truth to training point clouds while checking for collisions to prevent unrealistic outcomes.
We adopt a similar strategy for Anomaly Sample augmentation, constructing a database from SUN-RGBD data. From this database, we obtain anomaly samples using a copy-paste approach, treating them as an additional class (`Anomaly') for detector training. Anomaly Sample Augmentation trains the detector to localize unseen objects of various sizes or contexts.
Specifically, we directly utilize the database formed in previous research~\cite{qi2019deep} for indoor 3D object detection, which consists of 3D cuboids and their corresponding RGB-D point clouds.

\subsubsection{Learning of Universal Objectness}
Existing 3D object detectors often have a high correlation between classification scores and confidence scores. For instance, in a single-stage detector like SECOND, the confidence score $conf_{i}$ operates as $\max\{c_{1}, c_{2}, \dotsc, c_{k}\}$. However, we aim to enhance unseen object detection and OOD classification separately. Therefore, we propose the addition of a separate objectness node that is trained for decoupling these aspects.

\begin{equation} 
    \small
    \label{equation:focal_loss}
   FL(p_{t})= -\alpha(1-p_{t})^{\gamma}\log(p_{t}) , 
\end{equation}

We use the conventional Focal loss employed in RetinaNet~\cite{lin2017focal} with the established SECOND settings, setting $\alpha = 0.25$ and $\gamma = 2$. We label the foreground, including the ID class and the `Anomaly' class, as 1, and everything else as 0. The objectness loss constructed with Focal loss is denoted as $L_{obj}$.
% Our goal is to model a universal objectness that includes the additional 'Anomaly' class. This is similar to the Region Proposal Network's objectness node in Faster-RCNN~\cite{girshick2015fast}, which universally generates proposals and performs localization.
% Specifically, for a single-stage detector, this directly becomes the confidence score. For a two-stage detector, it functions as a bridge, forming proposals as utilized in Faster-RCNN. The final confidence score for the two-stage detector is obtained through the second-stage classifier.
The introduced objectness node aims to model a universal objectness, akin to Faster-RCNN~\cite{girshick2015fast}'s Region Proposal Network. In a single-stage detector, it serves as the confidence score, while in a two-stage detector, it acts as a bridge, forming proposals for subsequent stages. The final confidence score for the two-stage detector is derived through the second-stage classifier.

\subsubsection{Learning of Detecting Unseen Objects}
% sub-figure
% 2,4,5
% no aug가 오리지널

% \begin{figure*}[t!]
\begin{figure}[t!]
\centering

\begin{subfigure}{0.49\linewidth}
\includegraphics[width=1.0\linewidth]{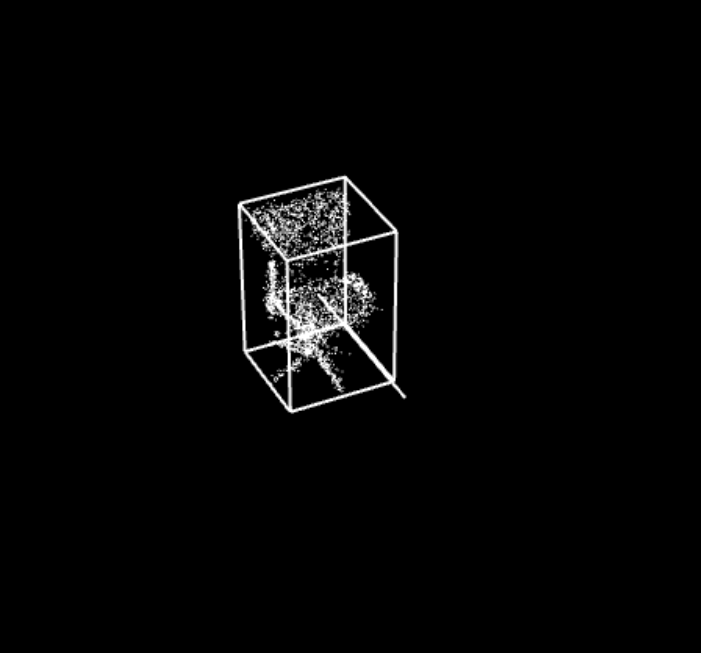}
% \caption{}
% \label{}
\end{subfigure}
\begin{subfigure}{0.49\linewidth}
\includegraphics[width=1.0\linewidth]{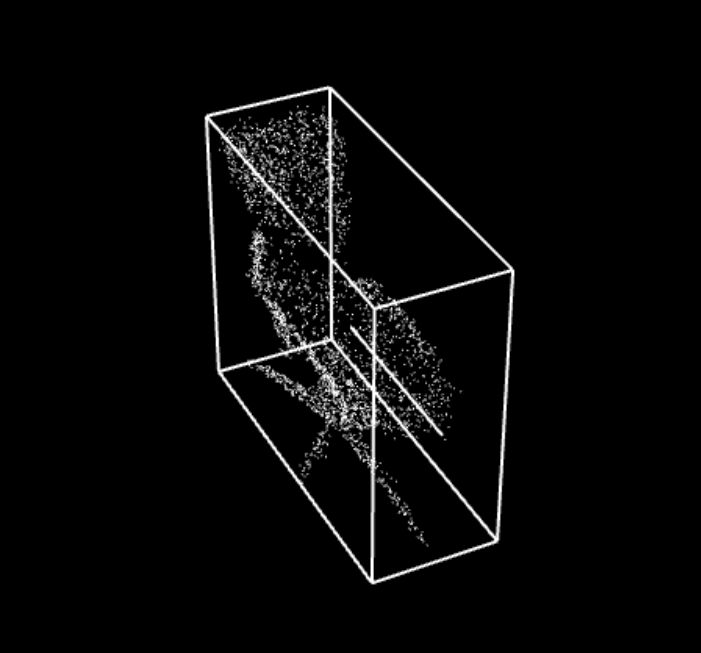}
% \caption{}
% \label{}
\end{subfigure}
\begin{subfigure}{0.49\linewidth}
\includegraphics[width=1.0\linewidth]{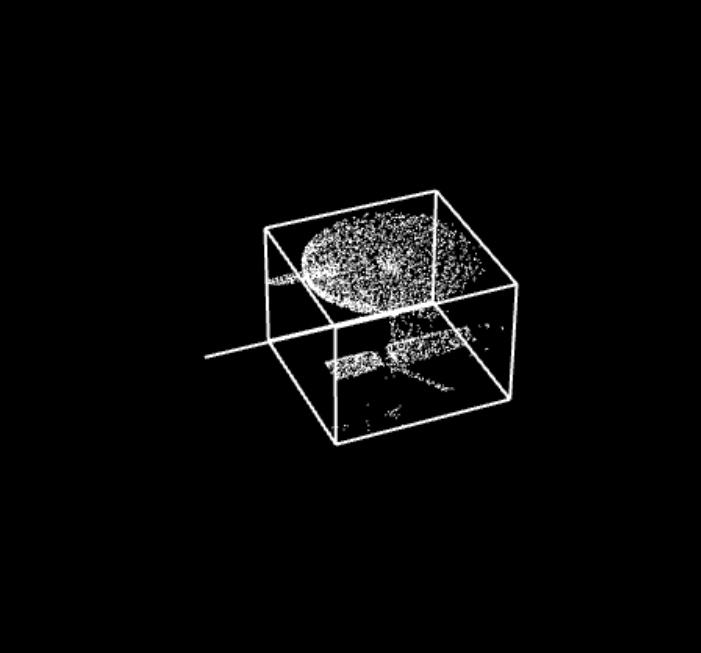}
\caption{Original}
\label{fig3_a}
% \caption{}
% \label{}
\end{subfigure}
\begin{subfigure}{0.49\linewidth}
\includegraphics[width=1.0\linewidth]{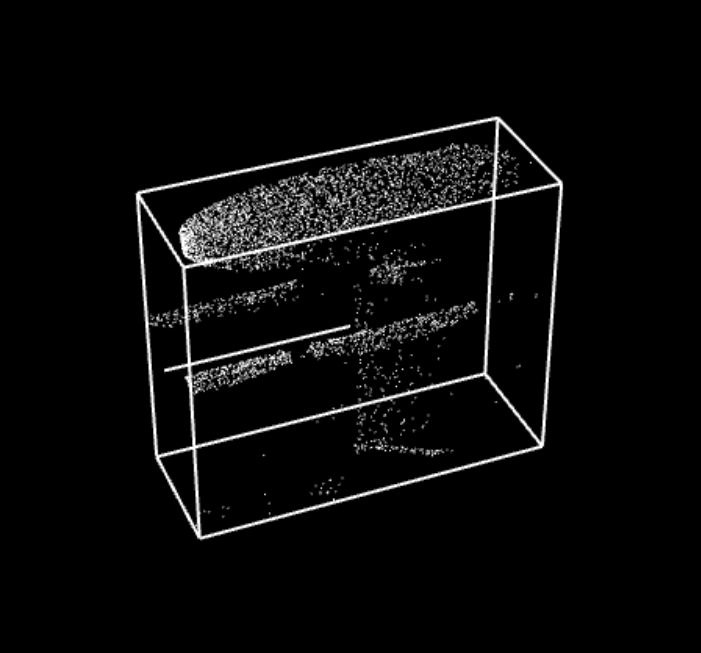}
\caption{Resized}
\label{fig3_b}

% \caption{}
% \label{}
\end{subfigure}
% \begin{subfigure}{0.49\linewidth}
% \includegraphics[width=1.0\linewidth]{figure/aug/no_aug5.pdf}
% \caption{Original}
% % \label{}
% \end{subfigure}
% \begin{subfigure}{0.49\linewidth}
% \includegraphics[width=1.0\linewidth]{figure/aug/aug5.pdf}
% \caption{Resized}
% % \label{}
% \end{subfigure}

\caption{\textbf{Visualization result on SUN-RGBD~\cite{song2015sun} pointcloud of original and resized object.} (a): Point cloud of the original object for Anomaly Sample Augmentation; (b): Point cloud of the resized object for Multi-size Mix Augmentation.}
% \label{}
\end{figure}

We train the model to detect objects of various sizes by adding the `Anomaly' class with Anomaly Sample augmentation. 
However, as shown in Figure~\ref{fig3_a}, the sizes of indoor scene data are generally smaller or less diverse compared to outdoor scenes. To address this, we propose Multi-size Mix augmentation to create a more diverse set of anomaly objects.
As illustrated in Figure~\ref{fig3_b}, we construct a database by resizing the original anomalies to various sizes and mixing them together. Specifically, Multi-size mix augmentation combines equal parts of the original anomaly at its original size and the resized anomaly. Additionally, the sizes for resizing the boxes are randomly extracted from various samples of box sizes in the KITTI Misc class.
% 하지만, 그림에서와 같이 indoor scene data 들의 size는 outdoor scene 에 비해 대체로 작거나 다양하지 못하다. 우리는 더 다양한 size의 anomaly 물체를 구성하기 위해 Multi-size mix augmentation 을 제안한다. 
% 그림 (b) 에서와 같이 기존의 anomaly 에 대해 다양한 size로 resize하여 이를 함께 mix 한 databse를 구성한다. 구체적으로, Multi-size mix augmentation 은 원래의 original size의 anomaly 와 resize 된 anomlay를 반반 섞어서 사용한다. 또한, resize 하는 box 의 크기로 단순하게 KITTI의 Misc class의 여러 샘플의 box size 로부터  random하게 추출하여 사용한다.

\subsubsection{Learning of Distinguishing Unseen Objects}
% one vs rest classifier 에 대해서 outlier data를 사용하는 것에 있어서 이전의 단순한 OE loss의 적용은 효과적이지 않다.
% 왜냐하면, 추가적인 outlier class를 설정하고 학습하는 기본적인 classifier 자체가기존의 in-distribtuion class에 대해 0으로 주도록 하고있다. 
% 따라서, 우리는 energy regularization loss~\cite{liu2020energy}를 통해 이를 해결하면서, 추가적으로 representation에서 ID-OOD 간의 seperability를 향상 시키는 outlier-aware contrastive learning~\cite{choi2023three}을 함께하도록 하는 것을 제안한다.
The straightforward application of the previous simple OE loss is not effective when using Anomaly data for a one-vs-rest classifier. This is because the basic classifier already trains an additional Anomaly class as it should go to all zero for the existing In-Distribution (ID)(seen) classes. Therefore, we address this issue by introducing energy regularization loss~\cite{liu2020energy}, 
Furthermore, we enhance performance by incorporating outlier-aware contrastive learning~\cite{choi2023three}, which improves the separability between In-Distribution (ID) and Out-Of-Distribution (OOD) data in the representation.
% submission할때 arxiv 논문을 ref하는건 최소화하는게 좋을 것 같은데 (특히 다른곳에서는 citation이 전혀 안된 우리 논문이기도 해서) 위의 ~\cite{choi2023three} citation은 삭제하거나 다른 논문 ref로 바꾸면 좋을것 같습니다.  
Energy regularization loss is defined by
    \begin{equation} 
    \small
    \label{equation:energy}
    \begin{split}
    &L_{en}=L_{in,hinge}+L_{out,hinge}
    \\ &=\mathbb{E}_{(\mathbf{x}_{in},y)\sim{D_{in}^{train}}}[(\max(0,E(\mathbf{x})-m_{in}))^{2}]
    \\ &+\mathbb{E}_{\mathbf{x} \sim{D_{out}^{train}}}[(\max(0,E(\mathbf{x})-m_{out}))^{2}].
\end{split}
\end{equation}
Here,  $D_{out}^{train}$ is defined as an 'Anomaly' class object.
where $E(\mathbf{x};f)=-T \cdot \log(\sum_{j=1}^{K}e^{f_{j}(\mathbf{x}))/T})$.
Energy function $E(\mathbf{x};f)$ is computed as LogSumExp of logit with temperature scaling,  In our cases, temperature $T$=$1$. Energy regularization loss is the sum of squared hinge losses for energy with each of the existing ID data and the auxiliary OOD (`Anomaly' class) data.   

The loss for contrastive learning is defined by
\begin{equation} 
    \small
    \label{equation:c_total_OOD}
   L_{c}= \sum_{i \in B_{in}}L_{i}, 
\end{equation}

\begin{equation} 
    \small
    \label{equation:c_each_OOD}
   L_{i}= -\frac{\mathbf{1}_{\{|B^{in}_{y_{i}}|>1\}}}{|B^{in}_{y_{i}}|-1} \sum_{p \in B^{in}_{y_{i}} \setminus \{i\}}{\log \frac{exp(\widetilde{\mathbf{f}}_{i}\cdot \widetilde{\mathbf{f}}_{p} / \tau_{c}) }{ \sum_{k \in  B^{all} \setminus \{i\}}{ exp (\widetilde{\mathbf{f}}_{i}\cdot \widetilde{\mathbf{f}}_{k} / \tau_{c})}}},
\end{equation} where we set $~ \frac{\mathbf{1}_{\{|B^{in}_{y_{i}}|>1\}}}{|B^{in}_{y_{i}}|-1}$ = $0$ when $|B^{in}_{y_{i}}|=1$;   \\ $\mathbf{1}_{\{|B^{in}_{y_{i}}|>1\}}$ = 1 when $|B^{in}_{y_{i}}|>1$. 

Within the total batch $B^{all}$, an instance $\mathbf{x}_{i}$ holds the following representation $\widetilde{\mathbf{f}}_{i}$.
$B^{all}$ has partition $B^{in}$ and $B^{out}$, each of which is an ID object and an Anomaly class object, respectively.  $B_{y_{i}}$ is a subset of the set $B$ where the label of every sample in $B_{y_{i}}$ matches $y_{i}$. Outlier-aware contrastive learning takes into account outlier data in addition to the traditional supervised contrastive learning (SCL)~\cite{khosla2020supervised}.
As a result, total loss $L_{total}$ for our loss 
is defined by
\begin{equation} 
    \small
    \label{equation:total}
L_{total}=L_{cls}+ L_{reg}+L_{obj}+\lambda_{en}L_{en}+\lambda_{c}L_{c}.
\end{equation}

\subsection{Additional Synthetic OOD Benchmark}

%\input{table_latex/C_algorithm}
% 1054

\begin{figure}[t!]
    \centering
    \includegraphics[width=1.0\linewidth]{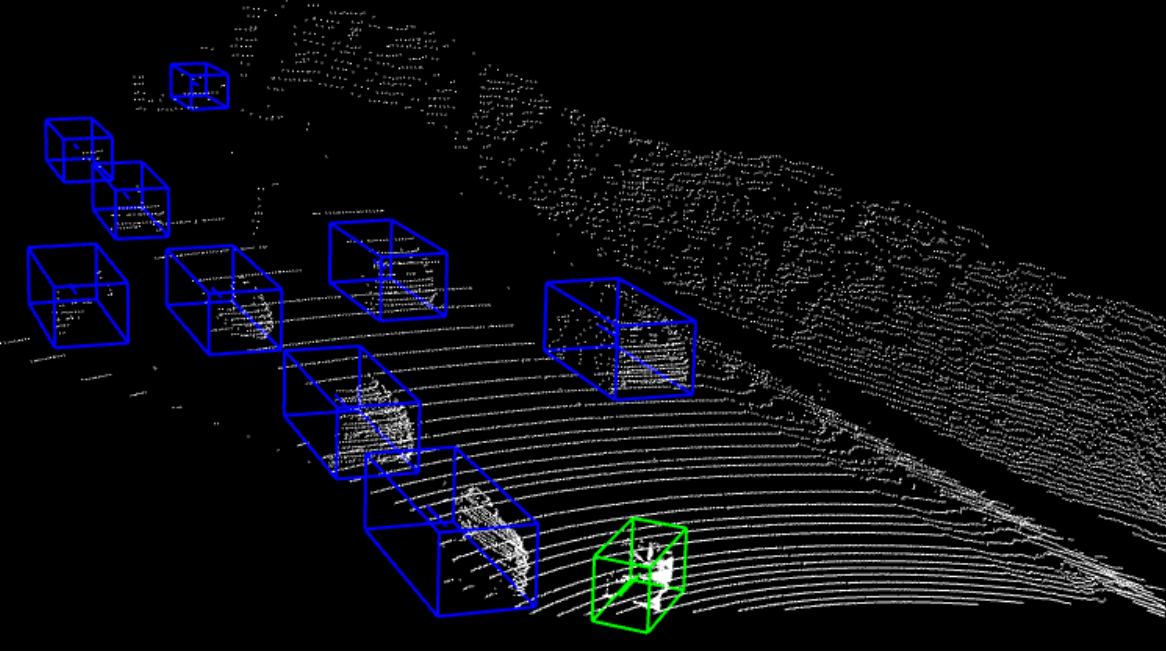}
    \caption{\textbf{Visualization on our proposed synthetic benchmark.} The blue box represents the original ID (seen) object, while the green box represents our cut-pasted synthesized OOD (unseen) object.} 
    % \label{num_samples}
    \label{syn_bench_fig}
\end{figure}
% 이전에 우리는 unseen object detection을  평가하기 위해 KITTI에서 'Misc' 클래스를 사용한 벤치마크를 제안하였습니다. 그러나 우리는 KITTI 'Misc' 보다 더 다양한 종류의 unseen object이 나타나는 시나리오를 만들기 위해, 외부의 데이터로부터 unseen object pointcloud instance 를 KITTI scene에 대해 인스턴스를 잘라내어 삽입하는 '잘라 붙이기' 기술을 통해 추가적으로 합성된 OOD benchmark를 제안합니다.
Aiming to create scenarios with a more diverse range of unseen objects than KITTI's 'Misc' class, we further propose two additional synthetic OOD benchmarks by inserting instances of unseen object point clouds from external data into KITTI scenes using the cut-paste technique. This enables us to incorporate a wider variety of unseen objects into the benchmark.
As depicted in Figure~\ref{syn_bench_fig}, our benchmark involves adding unseen objects to existing scenes. Blue represents the original in-distribution (seen) data, while green depicts the copy-pasted unseen objects. We aim to evaluate unseen object detection and OOD classification for existing baseline 3D detectors in scenes where these coexist.

Firstly, we propose the Nuscenes~\cite{caesar2020nuscenes} OOD benchmark by extracting unseen object point cloud instances from the large-scale outdoor dataset, Nuscenes. We select a total of five unseen object classes: Debris, Pushable Pullable Object, Traffic Barrier, Traffic Cone, and Animal, which correspond to objects not typically seen in KITTI scenes. For each class, we propose a total of five benchmarks each representing a different unseen object scenario.

Secondly, we propose the SUN-RGBD~\cite{song2015sun} OOD benchmark by extracting unseen object point cloud instances from the indoor dataset, SUN-RGBD. Objects in SUN-RGBD indoor scenes are naturally unseen in the outdoor KITTI dataset. We compose a set of unseen objects using the remaining five classes from SUN-RGBD that are not used for training. However, due to the relatively small number of object instances in SUN-RGBD point clouds, we propose a single benchmark scenario that uses all five classes. More detailed methods for constructing challenging benchmarks are described in the supplementary material.

\section{Experimental Result}
\subsection{Experiment Settings}
% KITTI training , validation 5 5 split set에서 실험을 진행한다.
% baseline 구성에 있어서 baseline detector는 기존 Open Pcdet 의 코드~\cite{openpcdet2020}를 기반으로 학습하였으며,
% 기존과 다른점은, training set에서 in-distribtuion class로 Car Pedestrian Cyclist 를 제외한 
% 나머지 class Person sitting , Truct 등등 은 background 로 학습이 되지 않기 위해 point cloud 에서 제거하고 학습한점이다.
% 우리는 일괄적으로 최대 500개의 detection result를 얻기 위해 
% 기존 OpenPcdet 에서 SECOND, PointPillar는 원래의 config setting 을 그대로 유지하되
% PV-RCNN와 PartA2에서는 first stage에서 proposal을 넘기는 부분에서 PRE MAX 8196 POST MAX 2048 으로 키워
% 최종적으로 최대 500개의 detection result를 얻도록 통일시켰다.

% Sun-RGBD의 processing 관련 (R,G,B,x,y,z)을 이용한다
% ~\cite{song2015sun} 에 대해 processing한 protocol 따른다~\cite{qi2019deep} R GB를 평균하여 intenstity로 변환
% (I,x,y,z) 4D vector로 변환 사용

% Misc bench mark에 대해서,  기존 validation set을 그대로 사용하되, 0~50m distance에 있는 Misc 가 있는 scene만을 선택하여 그 안에있는 coexist 하는 in-distribution sample 에 대해 수집하여 IND OOD 분포를 형성한다. 
% Recall 또한 마찬가지로, 이러한 scene에 대해서 OOD recall을 측정하였다

% 자세한 hyperparameter setting 및 학습환경은 supplementary에 자세하게 다룬다.

We conduct experiments on the KITTI~\cite{geiger2012we} training and validation sets with a 5:5 split. For the baseline configuration, the baseline detector is trained based on the code of OpenPCDet~\cite{openpcdet2020}. The key difference is that, in the training set, classes other than Car, Pedestrian, and Cyclist (e.g., Truck, Van, etc.) were removed from the point cloud to avoid training them as background. Also, we consistently aim to obtain a maximum of 500 detection results. For this purpose, SECOND and PointPillar maintain their original configuration settings from OpenPCDet. For PV-RCNN and PartA2, we changed the settings for inference in the first stage, increasing the NMS configuration of pre-max size to 8196 and post-max size to 2048 to ensure a lot of detection results.
We utilized the  $\{R, G, B, x, y, z\}$ information from the SUN-RGBD dataset and followed the processing protocol outlined in~\cite{song2015sun} and~\cite{qi2019deep}. The RGB values were averaged to convert them into intensity $\{I, x, y, z\}$, forming a 4D vector as same as KITTI.

For the KITTI Misc benchmark, we used the existing validation set but selected only scenes with Misc objects within the 0-50m distance range. From each scene where unseen objects coexist with seen objects, we collected OOD samples and ID samples for use in OOD classification. The recall was also measured by aggregating these scenes to evaluate unseen object recall. This is the same setting for a synthetic benchmark.
Detailed hyperparameter settings and training environments are described in the supplementary material.

% 우리의 방법 또한 기존 baseline 과 같이 나머지 class 제거 및 PRE MAX POST MAX는 그대로 동일하다.
% 다만, 다른점은 추가로 outlier class를 추가하여 학습하는 점이고 이들은 various size를 갖으므로, 추가적인 anchor box setting을 구성한다. anchor box setting은 다음과 같다.

% 우리의 방법을 위한 하이퍼파라미터 세팅은 supplementary에 자세하게 다룬다.
% contrastive loss
% loss parameter 
% energy loss parameter
% objectness loss parameter

% 우리의 방법을 위한 augmentation setting은 다음과 같다. 기존의 KITTI in-distribution 
% Car PEd Cyc 이것을 유지하면서 추가로 Min point number 5 
% training 에 있어서 cut-paste 하는 syn 의 개수는 20

% Sun-RGBD의 processing 관련 (R,G,B,x,y,z)을 이용한다
% ~\cite{song2015sun} 에 대해 processing한 protocol 따른다~\cite{qi2019deep} R GB를 평균하여 intenstity로 변환
% (I,x,y,z) 4D vector로 변환 사용

% contrastive learning을 위한 추가적인 network 
% feature의 수는 다음과 같다.

% Misc bench mark에 대해서,  기존 validation set을 그대로 사용하되, 0~50m distance에 있는 Misc 가 있는 scene만을 선택하여 그 안에있는 coexist 하는 in-distribution sample 에 대해 수집하여 IND OOD 분포를 형성한다. 
% Recall 또한 마찬가지로, 이러한 scene에 대해서 OOD recall을 측정하였다. 

% Synthetic benchmark 는 실제 training 에서와 같이 cut-paste augmentation을 하여 이에 대해 test하는 방식이다.
% 고정된 benchmark 구성을 위해 random seed를 1로 고정한다. 그리고 misc bechmkark 와 OOD sample수를 비슷하게 맞추기 위해
% misc 가 있는 scene 에 대해서만 합성하여 구성한다.  합성하는 syn의 개수 1로 설정하였다. 이 때, 기존 validation set 에서 misc를 포함한 기존의 나머지 foreground class들은 모두 제거하고 syn sample을 합성하여 넣었다.
% NN grid sampling을 위한 slice number $N=5$ 로 설정한다.
% Misc bench mark 와 비슷하게 합성된 결과 중 0~50m 에 해당하는 outlier가 존재하는 scene만을 선택하고 같은 setting을 따른다. 
% 학습을 위한 환경으로 우리는 batchsize 4 그리고 GPU8 개 환경에서 학습하였다.

\subsection{Evaluation on KITTI Misc benchmark}

\subsubsection{Quantitative Result}
% 우리의 방법은 KITTI 데이터에 대해서 검증을 하였으며, 특히 Unidenetified Foreground Object 로 Misc 클래스에 대해서 
% 대표적인 base line SECOND 에 비해 우수한 localization 성능을 보인다.
% 그래프에서 볼수있듯이, Proposal number 에 관계없이 그리고, IOU threshold 0.1, 0.25, 0.40 에서 모두 우수한 recall을 보여준다.
% 기존의 SECOND는 IOU threshold 0.40에서 9.28\%로 대부분의 Misc에 대해 정확한 localization에 실패하고 있다. 하지만, 우리의 방법은 이전에 비해 24.12\%로 큰 향상을 이룬다. 
\begin{figure}[t!]
    \centering
    \includegraphics[width=0.9\linewidth]{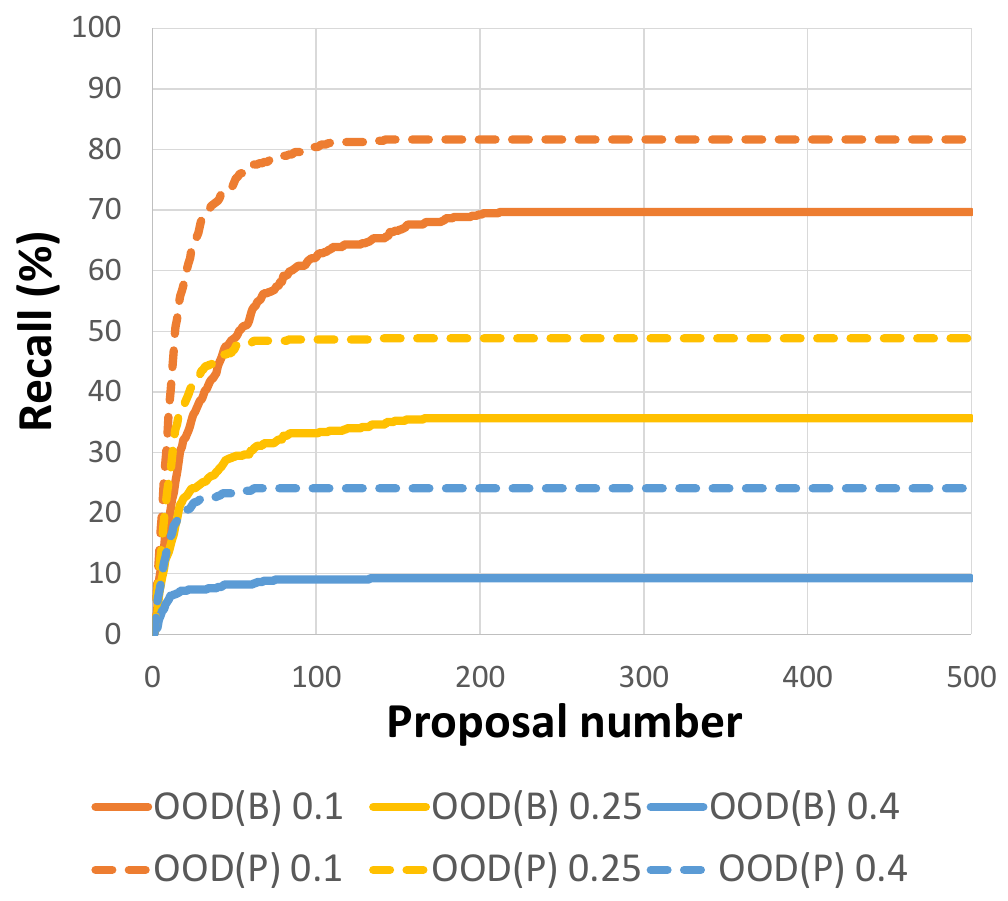}
    \caption{\textbf{OOD (unseen) object recall comparison on KITTI Misc benchmark}. OOD(B) represents the result of the baseline detector SECOND, and OOD(P) represents the result of our method on SECOND.}
    \label{comparsion_recall}
\end{figure}

% 다음으로 우리는 SECOND 를 넘어 4가지  PointPillars, PV-RCNN 및 Part-A2 네 가지 detector에 대해 Recall 및 OOD 성능을 평가하였다.
% 테이블에서 요약했듯이, 기존의 Single Stage detector 인 PointPillars 와 SECOND 에 비해 Two stage detector 인 PV-RCNN 및 Part-A2 가 OOD 성능 및 Recall 성능 모두에서 우수한 경향을 보인다. 특히, PV-RCNN은 OOD 성능 및 Recall 모두에서 제일 좋은 결과를 보인다. 고성능의 Two stage detector 가 
% Single stage detector에 비해 detection precision이 우수한데, 이와 비슷한 경향을 보이는 듯하다.
% 비슷하게, 기존 연구에서 Transformer 기반의 우수한 classifier가 OOD 성능에서도 우수한 성능을 보이는 것과 비슷한 경향을 보이는듯하다.

% 마지막으로, 우리의 방법은 4가지 detector 모두에 효과적으로 작용하여
% Recall 뿐만 아니라 OOD detection 성능을 모두 높인다.
% 이는 Intro figure에서 자세히 볼수있다. 그 결과 UFO 에 대해 우리의 방법은 기존 baseline 에 비해 잘 localization 하면서 Unknown으로 구분함으로 실제 자율주행 적용에 있어서 UFO 물체를 검출하여 알고리즘의 robustness를 확보한다.  

We first quantitatively validate our method on the KITTI Misc benchmark, particularly showcasing superior localization performance for the Misc class, compared to the prominent baseline, SECOND.
As depicted in Figure~\ref{comparsion_recall}, regardless of proposal number and IOU thresholds (0.1, 0.25, 0.40), our method consistently exhibits excellent recall. 
% In contrast, the conventional SECOND struggles with accurate localization, especially at an IOU threshold of 0.40, where it achieves only 9.28\% recall for Misc. In comparison, our method significantly improves this to 24.12\%.
\subsubsection{Comparison with baseline}

Our method goes beyond SECOND and evaluates unseen object detection (recall) and OOD classification performance for four detectors: SECOND, PointPillars, PV-RCNN, and Part-A2. As summarized in Table~\ref{A_main}, the two-stage detectors, PV-RCNN and Part-A2, outperform single-stage detectors (PointPillars and SECOND) in both unseen object detection and OOD classification. 
Our method significantly improves recall and OOD detection across all detectors, as shown in Figure~\ref{intro_fig} of Introduction. 

%%%%%%%%%%%%%%%%%%%%%%
% mean 이랑 std 모두 소수 셋째자리에서 반올림해서 둘째 자리까지만 표시했습니다.
% SC OOD benchmark 결과입니다.

\begin{table}[t!]
%\captionsetup{font=scriptsize}

\caption{Quantitative result of our method on KITTI Misc benchmark. 500 proposals are used for all cases.}
\centering
\scriptsize
\label{A_main}

\begin{tabular}{|cc|ccc|ccc|}
\hline
\multicolumn{2}{|c|}{\multirow{2}{*}{METHOD}}              & \multicolumn{3}{c|}{Recall @IOU}                                                           & \multicolumn{3}{c|}{OOD detection}                                                         \\ \cline{3-8} 
\multicolumn{2}{|c|}{}                                     & \multicolumn{1}{c|}{0.10}           & \multicolumn{1}{c|}{0.25}           & 0.40           & \multicolumn{1}{c|}{AUC$\uparrow$}            & \multicolumn{1}{c|}{AP$\uparrow$}             & FPR$\downarrow$            \\ \hline
\multicolumn{1}{|c|}{\multirow{2}{*}{SECOND}}       & Base & \multicolumn{1}{c|}{69.69}          & \multicolumn{1}{c|}{35.67}          & 9.28           & \multicolumn{1}{c|}{85.53}          & \multicolumn{1}{c|}{81.17}          & 78.14          \\
\multicolumn{1}{|c|}{}                              & \textbf{Ours} & \multicolumn{1}{c|}{\textbf{81.65}} & \multicolumn{1}{c|}{\textbf{48.87}} & \textbf{24.12} & \multicolumn{1}{c|}{\textbf{88.48}} & \multicolumn{1}{c|}{\textbf{82.94}} & \textbf{55.05} \\ \hline
\multicolumn{1}{|c|}{\multirow{2}{*}{PointPillars}} & Base & \multicolumn{1}{c|}{76.70}          & \multicolumn{1}{c|}{44.74}          & 18.14          & \multicolumn{1}{c|}{75.38}          & \multicolumn{1}{c|}{68.79}          & 89.48          \\
\multicolumn{1}{|c|}{}                              & \textbf{Ours} & \multicolumn{1}{c|}{\textbf{82.89}} & \multicolumn{1}{c|}{\textbf{54.85}} & \textbf{26.60} & \multicolumn{1}{c|}{\textbf{83.79}} & \multicolumn{1}{c|}{\textbf{76.76}} & \textbf{61.65} \\ \hline
\multicolumn{1}{|c|}{\multirow{2}{*}{PV-RCNN}}      & Base & \multicolumn{1}{c|}{85.98}          & \multicolumn{1}{c|}{54.43}          & 16.49          & \multicolumn{1}{c|}{86.28}          & \multicolumn{1}{c|}{80.79}          & 72.37          \\
\multicolumn{1}{|c|}{}                              & \textbf{Ours} & \multicolumn{1}{c|}{\textbf{89.48}} & \multicolumn{1}{c|}{\textbf{62.47}} & \textbf{31.13} & \multicolumn{1}{c|}{\textbf{90.43}} & \multicolumn{1}{c|}{\textbf{85.89}} & \textbf{40.21} \\ \hline
\multicolumn{1}{|c|}{\multirow{2}{*}{PartA2}}      & Base & \multicolumn{1}{c|}{80.21}          & \multicolumn{1}{c|}{44.54}          & 10.31          & \multicolumn{1}{c|}{85.63}          & \multicolumn{1}{c|}{79.73}          & 66.39          \\
\multicolumn{1}{|c|}{}                              & \textbf{Ours} & \multicolumn{1}{c|}{\textbf{88.87}} & \multicolumn{1}{c|}{\textbf{57.32}} & \textbf{23.09} & \multicolumn{1}{c|}{\textbf{88.45}} & \multicolumn{1}{c|}{\textbf{84.20}} & \textbf{46.39} \\ \hline
\end{tabular}
\end{table}

\subsubsection{Comparison with SOTA}

%%%%%%%%%%%%%%%%%%%%%%
% mean 이랑 std 모두 소수 셋째자리에서 반올림해서 둘째 자리까지만 표시했습니다.
% SC OOD benchmark 결과입니다.

\begin{table}[t!]
%\captionsetup{font=scriptsize}

\caption{Comparison result with SOTA on KITTI Misc benchmark. 500 proposals are used for all cases.}
\centering
\scriptsize
\label{S_main}

\begin{tabular}{|c|ccc|ccc|}
\hline
\multirow{2}{*}{Method} & \multicolumn{3}{c|}{Recall @IOU}                                                           & \multicolumn{3}{c|}{OOD detection}                                                         \\ \cline{2-7} 
                        & \multicolumn{1}{c|}{0.10}           & \multicolumn{1}{c|}{0.25}           & 0.40           & \multicolumn{1}{c|}{AUC}            & \multicolumn{1}{c|}{AP}             & FPR            \\
                        \hline
% ODIN~\cite{liang2017enhancing}             & \multicolumn{1}{c|}{69.69}          & \multicolumn{1}{c|}{35.67}          & 9.28           & \multicolumn{1}{c|}{\textbf{86.17}} & \multicolumn{1}{c|}{78.11}          & \textbf{49.48} \\
% MaxLogit~\cite{hendrycks2019scaling}         & \multicolumn{1}{c|}{69.69}          & \multicolumn{1}{c|}{35.67}          & 9.28           & \multicolumn{1}{c|}{85.54}          & \multicolumn{1}{c|}{\textbf{81.17}} & 78.14          \\
% Energy~\cite{liu2020energy}           & \multicolumn{1}{c|}{69.69}          & \multicolumn{1}{c|}{35.67}          & 9.28           & \multicolumn{1}{c|}{85.53}          & \multicolumn{1}{c|}{81.17}          & 78.14          \\
% JointEnergy~\cite{wang2021can}      & \multicolumn{1}{c|}{69.69}          & \multicolumn{1}{c|}{35.67}          & 9.28           & \multicolumn{1}{c|}{85.53}          & \multicolumn{1}{c|}{81.17}          & 78.14          \\
Huang et al.~\cite{huang2022out}             & \multicolumn{1}{c|}{76.70} & \multicolumn{1}{c|}{44.74} & 18.14 & \multicolumn{1}{c|}{70.55}          & \multicolumn{1}{c|}{62.37}          & 85.57          \\ \hline
\textbf{Ours} (SECOND)            & \multicolumn{1}{c|}{81.65}          & \multicolumn{1}{c|}{48.87}          & 26.60          & \multicolumn{1}{c|}{88.48}          & \multicolumn{1}{c|}{82.94}          & 55.05          \\
\textbf{Ours} (PointPillars)      & \multicolumn{1}{c|}{82.89}          & \multicolumn{1}{c|}{54.85}          & 26.60          & \multicolumn{1}{c|}{83.79}          & \multicolumn{1}{c|}{76.76}          & 61.65          \\
\textbf{Ours} (PV-RCNN)            & \multicolumn{1}{c|}{\textbf{89.48}} & \multicolumn{1}{c|}{\textbf{62.47}} & \textbf{31.13} & \multicolumn{1}{c|}{\textbf{90.43}} & \multicolumn{1}{c|}{\textbf{85.89}} & \textbf{40.21} \\
\textbf{Ours} (PartA2)            & \multicolumn{1}{c|}{88.87}          & \multicolumn{1}{c|}{57.32}          & 23.09          & \multicolumn{1}{c|}{88.45}          & \multicolumn{1}{c|}{84.20}          & 46.39          \\ \hline
\end{tabular}
\end{table}

To the best of our knowledge, \cite{huang2022out} is the only existing work on OOD classification for 3D object detector. Therefore, for a comparison with state-of-the-art (SOTA), we compare \cite{huang2022out} using MSP, which shows the best or competitive performance across all datasets. Following \cite{huang2022out}, we leverage a PointPillars detector for Huang et al.~\cite{huang2022out}. Table~\ref{S_main} demonstrates the comparison between \cite{huang2022out} and our method, in terms of metrics on both unseen object localization and OOD classification. Our method outperforms \cite{huang2022out} with large margins across all evaluation metrics, irrespective of the backbone network.

% 기존의 방법 중에 직접적으로 3D object detection 에서 OOD detection을 수행한
% Huang et al~\cite{huang2022out} 는 PointPillars detector에 대해 여러가지 방법을 비교하였으나
% 그 중에 대체로 가장 좋은 Max Softmax 방법과 비교하였다.
% 나머지 방법들은 SECOND detector 에 대해 post processing을 사용한 OOD 최신방법들로 
% ODIN, MaxLogit, Energy, JointEnergy 4가지 방법에 대해 구현하여 비교하였다.
% 대체로 ODIN 이외에는 대부분 post processing에 영향을 크게 받지 않았다.
% Table~\ref{S_main} 에서 보이듯이, 우리의 방법이 최신의 방법들과 비교하였을때 OOD detection 성능과 OOD object를 localizaiton함에 있어서 모두 우월한 성능을 보이고 있다

\subsubsection{Qualitative Result}
\begin{figure*}[t!]
    \centering
    \includegraphics[width=\linewidth]{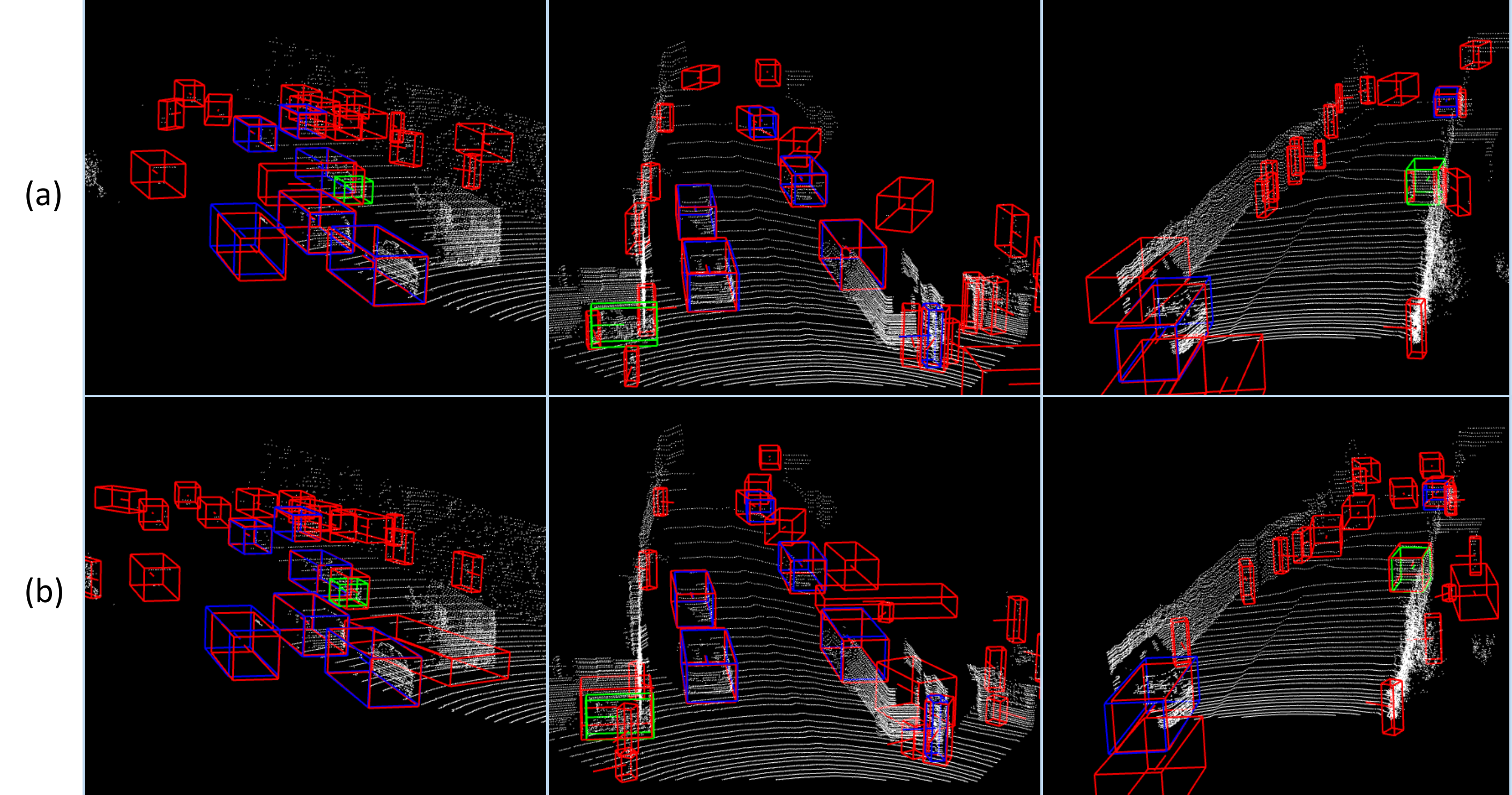}
    \caption{\textbf{Qualitative result of our method on KITTI Misc benchmark.} (a): Base detector result; (b): Our result.}
    % \label{num_samples}
    \label{fig_qualit}
\end{figure*}

We qualitatively validate our method through visualization, specifically against the baseline SECOND detector. As shown in Figure~\ref{fig_qualit}, the top images depict the results of the conventional SECOND, while the bottom images showcase our method. Blue boxes represent ground truth boxes for in-distribution, and green boxes represent ground truth boxes for Misc. The red boxes indicate the Top-25 results from the final detection.
In contrast to the baseline, which estimates unseen object 'Misc' localization with significantly different-sized boxes, our method consistently provides more accurate estimates with boxes of similar sizes. The superiority of our approach is visually evident, confirming its effectiveness.

% 우리의 방법에 대해 visualization 결과를 통해 
% 비교하고자 한다.
% 우리는 baseline SECOND detector 의 결과와 우리의 방법을 비교한다.
% 그림에서 나와있듯이, 위의 그림들은 기존 SECOND의 결과이고
% 아래의 그림은 우리의 방법이다. 파란색 박스는 Indistribution 에 대한 ground truth 박스이고, 초록색 박스는 Misc 에 대한 ground truth 박스이다.
% 빨간색 박스는 최종 detection 결과 중 top 25 result를 뽑아 도시한것이다.
% 기존의 baseline 은 Misc의 localization에 있어서 전혀 다른 size의 box로 추정을 하는 것에 비해 우리의 방법은 비슷한 size 의 box로 추정하는 것을 확인할수있다. 우리 방법의 우수성을 visualizaiton을 통해 확인한다.
\subsection{Evaluation on Nuscenes OOD benchmark}

\subsubsection{Comparison with baseline}
% \usepackage{multirow}
\begin{table}[t!]
%\captionsetup{font=scriptsize}

\caption{Quantitative result of our method (SECOND) on Nuscenes OOD benchmark.}
\centering
\scriptsize
\label{NUSC_SECOND}
\begin{tabular}{|c|c|ccc|ccc|}
\hline
\multirow{2}{*}{Class}       & Method        & \multicolumn{3}{c|}{Recall @IOU}                                                           & \multicolumn{3}{c|}{OOD detection}                                                         \\ \cline{2-8} 
                                          & SECOND        & \multicolumn{1}{c|}{0.10}           & \multicolumn{1}{c|}{0.25}           & 0.40           & \multicolumn{1}{c|}{AUC$\uparrow$}            & \multicolumn{1}{c|}{AP$\uparrow$}             & FPR$\downarrow$            \\ \hline
\multirow{2}{*}{Deb}                   & Base          & \multicolumn{1}{c|}{28.63}          & \multicolumn{1}{c|}{11.37}          & 3.14           & \multicolumn{1}{c|}{86.95}          & \multicolumn{1}{c|}{86.20}          & 65.69          \\
                                          & \textbf{Ours} & \multicolumn{1}{c|}{\textbf{43.61}} & \multicolumn{1}{c|}{\textbf{31.83}} & \textbf{14.54} & \multicolumn{1}{c|}{\textbf{95.41}} & \multicolumn{1}{c|}{\textbf{96.05}} & \textbf{28.29} \\ \hline
\multirow{2}{*}{PPO} & Base          & \multicolumn{1}{c|}{21.51}          & \multicolumn{1}{c|}{16.28}          & 7.17           & \multicolumn{1}{c|}{91.41}          & \multicolumn{1}{c|}{92.86}          & 55.23          \\
                                          & \textbf{Ours} & \multicolumn{1}{c|}{\textbf{26.59}} & \multicolumn{1}{c|}{\textbf{22.35}} & \textbf{12.91} & \multicolumn{1}{c|}{\textbf{95.04}} & \multicolumn{1}{c|}{\textbf{94.98}} & \textbf{31.98} \\ \hline
\multirow{2}{*}{TB}          & Base          & \multicolumn{1}{c|}{23.03}          & \multicolumn{1}{c|}{1.62}           & 0.40           & \multicolumn{1}{c|}{88.29}          & \multicolumn{1}{c|}{86.39}          & 59.39          \\
                                          & \textbf{Ours} & \multicolumn{1}{c|}{\textbf{36.49}} & \multicolumn{1}{c|}{\textbf{22.78}} & \textbf{14.31} & \multicolumn{1}{c|}{\textbf{96.13}} & \multicolumn{1}{c|}{\textbf{95.82}} & \textbf{21.77} \\ \hline
\multirow{2}{*}{TC}             & Base          & \multicolumn{1}{c|}{29.73}          & \multicolumn{1}{c|}{20.66}          & 7.92           & \multicolumn{1}{c|}{89.42}          & \multicolumn{1}{c|}{90.77}          & 63.51          \\
                                          & \textbf{Ours} & \multicolumn{1}{c|}{\textbf{36.80}} & \multicolumn{1}{c|}{\textbf{26.98}} & \textbf{11.37} & \multicolumn{1}{c|}{\textbf{94.79}} & \multicolumn{1}{c|}{\textbf{94.95}} & \textbf{38.54} \\ \hline
\multirow{2}{*}{Ani}                   & Base          & \multicolumn{1}{c|}{16.83}          & \multicolumn{1}{c|}{2.10}           & 0.00           & \multicolumn{1}{c|}{90.73}          & \multicolumn{1}{c|}{91.64}          & 61.38          \\
                                          & \textbf{Ours} & \multicolumn{1}{c|}{\textbf{38.00}} & \multicolumn{1}{c|}{\textbf{17.08}} & \textbf{4.03}  & \multicolumn{1}{c|}{\textbf{95.93}} & \multicolumn{1}{c|}{\textbf{95.55}} & \textbf{26.30} \\ \hline
\end{tabular}
\end{table}
We validate our method on the Nuscenes OOD benchmark, which comprises unseen objects in large-scale outdoor scenes not present in the outdoor KITTI dataset.
We constructed a total of five benchmarks by introducing five OOD objects from Nuscenes to the KITTI scene. We select five OOD (unseen) object classes: Debris (Deb), Pushable Pullable Object (PPO), Traffic Barrier (TB), Traffic Cone (TC), and Animal (Ani). These objects are nearly unseen objects in KITTI scenes. We introduce these OOD (unseen) objects on KITTI scenes by copy-pasting the point cloud data in the Nuscenes dataset.
As shown in Table~\ref{NUSC_SECOND}, our method on SECOND (SEC) outperformed the baseline in all five benchmarks. Similar results are obtained for PointPillars, PV-RCNN, and PartA2. We include all results in the supplementary.

\subsection{Evaluation on SUN-RGBD OOD benchmark}

\subsubsection{Comparison with baseline}
We validate our method on the SUN-RGBD OOD benchmark, which consists of unseen objects in indoor scenes not present in the outdoor KITTI dataset (also, as a training sample).
As summarized in Table~\ref{B_main}, conventional base detectors struggle with detecting unseen objects that are synthetically introduced from indoor scenes.
Consistent with the results from the Misc benchmark, high-performance two-stage detectors outperform single-stage detectors in OOD classification. Furthermore, applying our method to all four detectors substantially improves unseen object detection and enhances OOD classification performance. This trend holds across all detectors.
% 우리가 제안한 synthetic benchmark에서도 4가지 base detector 에 대해
% baseline을 구성하였으며, OOD detection 성능 및 Recall을 함께 도시하였다.
% 그림에서 요약하였듯이, 기존 base detector 들은 synthetic 하게 합성된 indoor-scene 에서의 물체들에 대해 localization 에 있어서 어려움을 겪는다.
% 하지만, Misc benchmark 에서와 같이 성능 좋은 two stage detector가 single stage detector 에 비해 OOD detection 성능은 우수한 경향을 보인다.
% 또한, 4가지 detector 모두에 있어서 우리의방법을 적용하면, Localizaiton 에 있어서 대폭 향상되며, OOD detection 성능에서도 향상되는 Misc 에서와 같이비슷한 경향을 가진다.
%\input{figure_latex/syn_result}
% Notably, our approach exhibits significant improvements in OOD classification, particularly when compared to the Misc benchmark. This pronounced enhancement can be attributed to the use of indoor scene data in the Anomaly Sample augmentation, which, despite having different classes, shares the same domain as the unseen object data. This makes OOD detection more straightforward compared to the outdoor scene with a Misc class object.
% 자세한 결과는 테이블에 요약하였으며, 우리의 방법은 역시 기존 baseline에 비해 우수한 성능을 보인다. 기존 Misc 에 비해서 특히 OOD detection 에 있어서 큰 성능 향을 보이는데, 우리가 Anomaly Mix augmentation으로 사용한 in-door scene data 와 class가 다르지만 같은 domain을 갖는  OOD data 이므로 OOD 로 판단함에 있어서  Out door scene 인 Misc 에 비해 수월한 것으로 판단된다.

% cifar 10 (q=0.75) / 10 LT (q=0.8) 는 one step
% cifar 100 (q=0.85) / 100 LT (q=0.9) 는 no step

\begin{table}[t!]
%\captionsetup{font=scriptsize}

\caption{Quantitative result of our method on SUN-RGBD OOD benchmark.}
\centering
\scriptsize
\label{B_main}

\begin{tabular}{|cc|ccc|ccc|}
\hline
\multicolumn{2}{|c|}{\multirow{2}{*}{METHOD}}                       & \multicolumn{3}{c|}{Recall @IOU}                                                           & \multicolumn{3}{c|}{OOD detection}                                                         \\ \cline{3-8} 
\multicolumn{2}{|c|}{}                                              & \multicolumn{1}{c|}{0.10}           & \multicolumn{1}{c|}{0.25}           & 0.40           & \multicolumn{1}{c|}{AUC$\uparrow$}            & \multicolumn{1}{c|}{AP$\uparrow$}             & FPR $\downarrow$           \\ \hline
\multicolumn{1}{|c|}{\multirow{2}{*}{SECOND}}       & Base          & \multicolumn{1}{c|}{69.40}          & \multicolumn{1}{c|}{25.28}          & 2.44           & \multicolumn{1}{c|}{84.63}          & \multicolumn{1}{c|}{85.99}          & 85.81          \\
\multicolumn{1}{|c|}{}                              & \textbf{Ours} & \multicolumn{1}{c|}{\textbf{92.36}} & \multicolumn{1}{c|}{\textbf{66.97}} & \textbf{34.16} & \multicolumn{1}{c|}{\textbf{96.94}} & \multicolumn{1}{c|}{\textbf{96.96}} & \textbf{9.44}  \\ \hline
\multicolumn{1}{|c|}{\multirow{2}{*}{PointPillars}} & Base          & \multicolumn{1}{c|}{67.69}          & \multicolumn{1}{c|}{22.49}          & 3.93           & \multicolumn{1}{c|}{76.23}          & \multicolumn{1}{c|}{74.49}          & 93.65          \\
\multicolumn{1}{|c|}{}                              & \textbf{Ours} & \multicolumn{1}{c|}{\textbf{85.56}} & \multicolumn{1}{c|}{\textbf{47.63}} & \textbf{23.49} & \multicolumn{1}{c|}{\textbf{90.28}} & \multicolumn{1}{c|}{\textbf{87.46}} & \textbf{32.97} \\ \hline
\multicolumn{1}{|c|}{\multirow{2}{*}{PV-RCNN}}      & Base          & \multicolumn{1}{c|}{80.27}          & \multicolumn{1}{c|}{24.22}          & 2.02           & \multicolumn{1}{c|}{90.58}          & \multicolumn{1}{c|}{88.62}          & 50.67          \\
\multicolumn{1}{|c|}{}                              & \textbf{Ours} & \multicolumn{1}{c|}{\textbf{97.33}} & \multicolumn{1}{c|}{\textbf{79.33}} & \textbf{43.56} & \multicolumn{1}{c|}{\textbf{96.25}} & \multicolumn{1}{c|}{\textbf{96.55}} & \textbf{5.12}  \\ \hline
\multicolumn{1}{|c|}{\multirow{2}{*}{PartA2}}      & Base          & \multicolumn{1}{c|}{73.97}          & \multicolumn{1}{c|}{21.91}          & 1.95           & \multicolumn{1}{c|}{87.37}          & \multicolumn{1}{c|}{84.40}          & 57.92          \\
\multicolumn{1}{|c|}{}                              & \textbf{Ours} & \multicolumn{1}{c|}{\textbf{96.05}} & \multicolumn{1}{c|}{\textbf{70.18}} & \textbf{37.06} & \multicolumn{1}{c|}{\textbf{96.33}} & \multicolumn{1}{c|}{\textbf{97.55}} & \textbf{5.70}  \\ \hline
\end{tabular}

\end{table}

\section{Discussion}
\subsection{Effect of objectness node}
\label{objectness_dis}

We train the classification score used in object detection and the objectness score used in localization separately. In the inference phase, comparing our objectness score with the traditional confidence score, as summarized in Table~\ref{C_ablation_objectness}, which confirms that our method achieves better performance in terms of localization.

\begin{table}[h!]
% \captionsetup{font=scriptsize}

\caption{Effect of using objectness node.}
\centering
\scriptsize
\label{C_ablation_objectness}
\begin{tabular}{|c|c|ccc|}
\hline
\multirow{2}{*}{METHOD}        & \multirow{2}{*}{Objectness node} & \multicolumn{3}{c|}{Recall @IOU}                                                        \\ \cline{3-5} 
                               &                                  & \multicolumn{1}{c|}{0.10}          & \multicolumn{1}{c|}{0.25}          & 0.40          \\ \hline
\multirow{2}{*}{\textbf{Ours} (SECOND)} & \xmark                              & \multicolumn{1}{c|}{78.97}         & \multicolumn{1}{c|}{47.01}         & 22.06         \\
                               & \cmark                              & \multicolumn{1}{c|}{\textbf{81.65}} & \multicolumn{1}{c|}{\textbf{48.87}} & \textbf{24.12} \\ \hline
\end{tabular}

\end{table}

\subsection{Comparison of OOD score metric}
% 우리는 모든 baseline 에 대해 Energy score metric으로 OOD 성능을 얻었다. 
% 기존 baseline 에 대해 기존의 다양한 score metric에 대해 얻은 AUROC 결과를 테이블에 정리하였다. 3D object detector에 있어서 OOD score metric에 따른 영향이 크지 않음을 확인할수있고, Energy score는 best는 아니지만 detector에 관계없이 안정적인 OOD 성능을 보여줌을 확인할수있다. 
We obtain OOD classification performance for all baselines using the Energy score metric. Table~\ref{C_score_result} summarizes AUROC results obtained for various score metrics on the existing baseline. It can be observed that the choice of OOD score metric has a limited impact on 3D object detectors. The Energy score, while not necessarily the best, consistently demonstrates stable OOD performance across detectors.

\begin{table}[h!]
% \captionsetup{font=scriptsize}

\caption{Comparsion of OOD score metric.}
\centering
\scriptsize
\label{C_score_result}
\begin{tabular}{|c|cccc|}
\hline
\multirow{2}{*}{Metric} & \multicolumn{4}{c|}{Method (AUROC$\uparrow$)}                                                                                              \\ \cline{2-5} 
                        & \multicolumn{1}{c|}{SECOND}         & \multicolumn{1}{c|}{PointPillars}   & \multicolumn{1}{c|}{PV-RCNN}        & Part-A2        \\ \hline
Max Logit~\cite{hendrycks2019scaling}               & \multicolumn{1}{c|}{85.54}          & \multicolumn{1}{c|}{75.37}          & \multicolumn{1}{c|}{86.28}          & 85.66          \\ 
Sum Logit~\cite{wang2021can}               & \multicolumn{1}{c|}{85.65}          & \multicolumn{1}{c|}{\textbf{76.07}} & \multicolumn{1}{c|}{\textbf{86.36}} & 83.56          \\ 
Max Prob~\cite{wang2021can}                & \multicolumn{1}{c|}{85.54}          & \multicolumn{1}{c|}{75.37}          & \multicolumn{1}{c|}{86.28}          & 85.66          \\ 
Sum Prob~\cite{wang2021can}                & \multicolumn{1}{c|}{85.53}          & \multicolumn{1}{c|}{75.38}          & \multicolumn{1}{c|}{86.28}          & 85.63          \\ 
MSP~\cite{hendrycks2016baseline}                     & \multicolumn{1}{c|}{\textbf{86.14}} & \multicolumn{1}{c|}{70.55}          & \multicolumn{1}{c|}{85.52}          & \textbf{86.20} \\ 
Max Energy~\cite{wang2021can}               & \multicolumn{1}{c|}{85.54}          & \multicolumn{1}{c|}{75.37}          & \multicolumn{1}{c|}{86.28}          & 85.66          \\ 
JointEnergy~\cite{wang2021can}               & \multicolumn{1}{c|}{85.53}          & \multicolumn{1}{c|}{75.38}          & \multicolumn{1}{c|}{86.28}          & 85.63          \\ \hline
Energy~\cite{liu2020energy}                  & \multicolumn{1}{c|}{85.53}          & \multicolumn{1}{c|}{75.38}          & \multicolumn{1}{c|}{86.28}          & 85.63          \\ \hline
\end{tabular}

\end{table}

\subsection{Ablation study on augmentation method}
\label{augment_dis}

% 우리는 in door scene으로 부터 얻은 Anomaly를 mix하는 augmentation 과 함께 multi-size mix augmentation을 사용함으로 Localization 성능을 크게 향상 시킨다.
% 표에 정리되었듯이 두 가지 augmentation을 모두 사용할때 제일 좋은 localization 성능을 얻음을 확인한다.
We significantly improve unseen object detection performance by employing Multi-size Mix (MM) augmentation in conjunction with the Anomaly Sample (AS) augmentation obtained from indoor scenes. As summarized in Table~\ref{D_ablation_multi_augment}, the combination of both augmentations yields the best-unseen object detection performance.

\begin{table}[h!]
% \captionsetup{font=scriptsize}

\caption{Augmentation method ablation result.}
\centering
\scriptsize
\label{D_ablation_multi_augment}
\begin{tabular}{|c|cc|ccc|}
\hline
\multirow{2}{*}{METHOD} & \multicolumn{2}{c|}{Augmentation method} & \multicolumn{3}{c|}{Recall @IOU}                                \\ \cline{2-6} 
                        & AS aug.      & MM aug.      & \multicolumn{1}{c|}{0.10}  & \multicolumn{1}{c|}{0.25}  & 0.40  \\ \hline
\multirow{3}{*}{\textbf{Ours} (SECOND)} & \xmark & \xmark                  & \multicolumn{1}{c|}{69.69} & \multicolumn{1}{c|}{35.67} & 9.28  \\
                        & \cmark                & \xmark                   & \multicolumn{1}{c|}{72.16} & \multicolumn{1}{c|}{40.62} & 17.11 \\
                        & \cmark                  & \cmark                  & \multicolumn{1}{c|}{\textbf{81.65}} & \multicolumn{1}{c|}{\textbf{48.87}} & \textbf{24.12} \\ \hline
\end{tabular}
\vspace{-0.2cm}
\end{table}

\subsection{Ablation study on loss}
% 우리는 OOD detection 성능 향상을 위한 추가적인 loss 로 energy loss 와 contrastive loss를 사용하였다. 테이블에 정리되었듯이, contrastive loss 사용시에 feature emebedding 에 있어서 ID-OOD 간의 seperability 가 좋아져 기존에 비해 OOD 성능을 크게 향상시킬수있음을 확인한다.
To enhance OOD classification performance, we incorporate additional losses, namely energy loss and contrastive loss. As summarized in Table~\ref{E_ablation_loss}, the use of contrastive loss significantly improves the separability between ID and OOD objects in feature embeddings, leading to a substantial enhancement in OOD classification performance compared to conventional methods.

\begin{table}[h!]
% \captionsetup{font=scriptsize}

\caption{Loss component ablation result.}
\centering
\scriptsize
\label{E_ablation_loss}
\begin{tabular}{|c|cc|ccc|}
\hline
\multirow{2}{*}{METHOD} & \multicolumn{2}{c|}{Loss component}       & \multicolumn{3}{c|}{OOD detection}                                                         \\ \cline{2-6} 
                        & Energy & \multicolumn{1}{l|}{Contrastive} & \multicolumn{1}{c|}{AUC$\uparrow$}            & \multicolumn{1}{c|}{AP$\uparrow$}             & FPR$\downarrow$            \\ \hline
\multirow{3}{*}{\textbf{Ours} (SECOND)} & \xmark      & \xmark                                & \multicolumn{1}{c|}{85.53}          & \multicolumn{1}{c|}{81.17}          & 78.14          \\
                        & \cmark      & \xmark                                & \multicolumn{1}{c|}{86.38}          & \multicolumn{1}{c|}{79.13}          & 58.35          \\
                        & \cmark      & \cmark                                & \multicolumn{1}{c|}{\textbf{88.48}} & \multicolumn{1}{c|}{\textbf{82.94}} & \textbf{55.05} \\ \hline
\end{tabular}
\vspace{-0.4cm}
\end{table}

\section{Conclusion}
% We proposed a novel protocol for assessing UFO detection on KITTI scenes, establishing baselines for four 3D object detectors: SECOND, PointPillars, PV-RCNN, and Part-A2.
% Our practical techniques significantly improve UFO detection in both localization and OOD detection compared to existing 3D object detector baselines. We create a new synthetic benchmark to model a diverse range of UFOs, validating our evaluation protocol and offering insights for future work on UFO detection in real-world scenarios. We believe our work suggests a future direction of 3D OOD detection toward enhancing localization performance as well as OOD distinction performance.

We introduce an integrated protocol for evaluating open-world 3D object detection on KITTI scenes, providing baseline assessments for four 3D object detectors: SECOND, PointPillars, PV-RCNN, and PartA2. By applying practical techniques to enhance the performance of existing 3D object detectors, we improve open-world 3D object detection performance in both unseen object detection and OOD classification. Additionally, by constructing three benchmark scenarios to model a diverse range of unseen objects, we confirm significant improvements in our method over baselines. We believe our work suggests a future direction for 3D object detectors in open-world scenarios, focusing on enhancing the detection and classification of unseen objects.

\clearpage

\setcounter{page}{1}
\maketitlesupplementary

\setcounter{equation}{5}
\setcounter{table}{8}
\setcounter{figure}{6}
%%%
%사회적 영향: 저희 작업의 넓은 영향과 한계를 심도 있게 탐구하며 사회적 영향을 논의합니다.

%데이터셋: 우리의 OOD 테스트 벤치마크에 대한 자세한 설명으로, 그 구성과 특성을 상세히 소개합니다.

%구현 세부 사항: 세부적인 구현 과정을 다루며, 훈련, 하이퍼파라미터 튜닝, 저희 방법 및 기타 방법의 구현 세부 사항을 자세히 설명합니다.

%시각화: 어려운 샘플 시각화의 자세한 묘사와 해석을 통해 모델 예측 해석에서의 중요성을 강조합니다.

%상세한 실험 결과: 각 OOD 테스트 데이터셋에 대한 결과의 종합적인 제시와 모든 인-분포 데이터셋에 대한 표준 편차가 포함된 철저한 벤치마크 비교 테이블을 제공합니다.

%네트워크 분석: WideResNet의 네트워크 분석의 세부 사항을 다루며, SC-OOD 벤치마크에 대한 표준 편차가 포함된 종합적인 비교 테이블을 제공합니다

Our supplementary material covers the following topics:

% \noindent \textbf{Societal Impact}: In-depth exploration of our work's broader impacts and limitations, addressing its societal impact.

\noindent \textbf {Algorithm Details for OOD Classification Evaluation}: Algorithm 1

\noindent \textbf {Nuscenes OOD Benchmark Full Result}:Table 9 and 10

\noindent \textbf {Implementation Details}: Specifics of our implementation process, including detailed training, hyperparameter tuning, and implementation details for baseline and our method.
%본문 내 공간 제한으로, 저희는 샘플의 난이도를 시각화한 결과와 그에 따른 해석을 다루며, 매우 어려운 샘플이 가져오는 현상에 대한 분석을 진행합니다.

\noindent \textbf {Original 3D Object Detection Results}: 
Due to spatial constraints in the main text, we provide a result of the original 3D object detection performance. Specifically, we present a comparison of the 3D object detection performance between the baseline model and our algorithm.

\noindent \textbf {More Visualization of Result on KITTI Misc Benchmark}: To supplement what couldn't be covered in the main text due to space limitations, we present additional visualization results for the KITTI Misc benchmark.

\noindent \textbf {Visualization of Result on Synthetic OOD Benchmark}: To supplement what couldn't be covered in the main text due to space limitations, we provide visualization results for the synthetic benchmark.

\noindent \textbf {Limitations}: We qualitatively analyze the limitations of our approach through visualization. Specifically, we compare and analyze failure cases of our method of localizing unseen objects in the KITTI Misc benchmark with the baseline.

\section{Algorithm Details for OOD Classification Evaluation}

We perform Out-Of-Distribution (OOD) classification based on scalar scores obtained from the In-Distribution (ID) classification score and OOD classification score from the final detection results. The evaluation metrics include AUROC, FPR95, and AUPR. 
In general, when ground truth and detection results have overlap, we can match them to obtain classification scores for both ID and OOD. However, in practice, it is common to encounter situations where there is no overlap for OOD data.
Therefore, for precise OOD classification evaluation, we propose a separate handling for such ground truth samples. We address cases where IOU is not available by matching the closest detection result based on Euclidean distance.
Our method follows the existing Hungarian-based matching, providing the advantage of avoiding heuristic parameter selection.
As outlined in Algorithm~\ref{algo1}, we first distinguish samples with no IOU and then handle them separately. For these cases, we perform matching based on distance to find the closest sample.
\SetKwInput{KwInput}{Input}                % Set the Input
\SetKwInput{KwOutput}{Output}   
% set the Output
\SetKwInput{Kwzerostep}{Step0}
\SetKwInput{Kwonestep}{Step1}
\SetKwInput{Kwtwostep}{Step2}
\begin{algorithm}[t]
\algsetup{linenosize=\tiny}
\footnotesize %
%\SetAlgoLined
\KwInput{$G_{i}$:  Ground truth $M$, $O_{j}$: Detection results $N$}

\KwOutput{$M_{i}$: Matching index result }
\Kwzerostep{
\textbf{Classify into results with overlap or no }}
 Get IOU matrix $IOU_{i,j} \leftarrow $ IOU between pairs ($G_{i}$,$O_{j}$) \\
 \For{$i=1$ to $M$} 
 {
  \uIf{ $IOU_{i,j}$ is all zero}{
    Gather as $A_{i}$
  }
  \Else{
    Gather as $B_{i}$
  }
 }
\Kwonestep{
\textbf{IOU based hungarian matching }}
Get IOU matrix $IOU_{i,j} \leftarrow $ IOU between pairs ($B_{i}$,$G_{j}$)  \\
Hungarian Matching $IM_{i}$ which maximize $IOU_{i,j}$\\
Remove matched result $C_{j}\leftarrow G_{j} $ \\
\Kwtwostep{
\textbf{Distance based hungarian matching}}
Get distance matrix $DIST_{i,j} \leftarrow $ IOU between pairs ($A_{i}$,$C_{j}$) \\
Hungarian Matching $DM_{i}$ which minimize $DIST_{i,j}$ \\

Aggregate and get final matching result 
$M_{i}$ $\leftarrow IM_{i}, DM_{i} $

\caption{Hungarian Based Matching }
\label{algo1}
\end{algorithm}

\section{Nuscenes OOD Benchmark Detailed Result}

We validate our method on the Nuscenes OOD benchmark, which comprises unseen objects in large-scale outdoor scenes not present in the outdoor KITTI dataset. For detailed results, we show the comparison results for other detectors: PointPillars and PartA2.
As shown in Table~\ref{NUSC_PP} and \ref{NUSC_Part}, our method on PointPillars and PartA2 also outperformed the baseline in all five benchmarks.

% \usepackage{multirow}
\begin{table}[t!]
%\captionsetup{font=scriptsize}

\caption{Quantitative result of our method (PointPillars) on Nuscenes OOD benchmark.}
\centering
\scriptsize
\label{NUSC_PP}
\begin{tabular}{|c|c|ccc|ccc|}
\hline
\multirow{2}{*}{Class}       & Method        & \multicolumn{3}{c|}{Recall @IOU}                                                           & \multicolumn{3}{c|}{OOD detection}                                                         \\ \cline{2-8} 
                                          & PointPillars  & \multicolumn{1}{c|}{0.10}           & \multicolumn{1}{c|}{0.25}           & 0.40           & \multicolumn{1}{c|}{AUC$\uparrow$}            & \multicolumn{1}{c|}{AP$\uparrow$}             & FPR$\downarrow$            \\ \hline
\multirow{2}{*}{Deb}                   & Base          & \multicolumn{1}{c|}{31.51}          & \multicolumn{1}{c|}{17.61}          & 4.89           & \multicolumn{1}{c|}{82.30}          & \multicolumn{1}{c|}{78.31}          & 69.67          \\
                                          & \textbf{Ours} & \multicolumn{1}{c|}{\textbf{38.60}} & \multicolumn{1}{c|}{\textbf{29.43}} & \textbf{15.20} & \multicolumn{1}{c|}{\textbf{90.50}} & \multicolumn{1}{c|}{\textbf{88.45}} & \textbf{43.86} \\ \hline
\multirow{2}{*}{PPO} & Base          & \multicolumn{1}{c|}{22.61}          & \multicolumn{1}{c|}{17.24}          & 8.24           & \multicolumn{1}{c|}{85.06}          & \multicolumn{1}{c|}{80.37}          & 64.37          \\
                                          & \textbf{Ours} & \multicolumn{1}{c|}{\textbf{25.92}} & \multicolumn{1}{c|}{\textbf{22.24}} & \textbf{11.99} & \multicolumn{1}{c|}{\textbf{88.11}} & \multicolumn{1}{c|}{\textbf{85.67}} & \textbf{55.32} \\ \hline
\multirow{2}{*}{TB}          & Base          & \multicolumn{1}{c|}{17.76}          & \multicolumn{1}{c|}{2.00}           & 0.40           & \multicolumn{1}{c|}{85.17}          & \multicolumn{1}{c|}{79.73}          & 64.87          \\
                                          & \textbf{Ours} & \multicolumn{1}{c|}{\textbf{27.33}} & \multicolumn{1}{c|}{\textbf{11.54}} & \textbf{8.50}  & \multicolumn{1}{c|}{\textbf{89.81}} & \multicolumn{1}{c|}{\textbf{84.42}} & \textbf{47.37} \\ \hline
\multirow{2}{*}{TC}             & Base          & \multicolumn{1}{c|}{28.57}          & \multicolumn{1}{c|}{17.76}          & 5.21           & \multicolumn{1}{c|}{87.57}          & \multicolumn{1}{c|}{89.39}          & 69.88          \\
                                          & \textbf{Ours} & \multicolumn{1}{c|}{\textbf{34.56}} & \multicolumn{1}{c|}{\textbf{22.59}} & \textbf{7.14}  & \multicolumn{1}{c|}{\textbf{90.49}} & \multicolumn{1}{c|}{\textbf{89.52}} & \textbf{51.16} \\ \hline
\multirow{2}{*}{Ani}                   & Base          & \multicolumn{1}{c|}{15.62}          & \multicolumn{1}{c|}{2.48}           & 0.00           & \multicolumn{1}{c|}{87.21}          & \multicolumn{1}{c|}{86.60}          & 67.62          \\
                                          & \textbf{Ours} & \multicolumn{1}{c|}{\textbf{30.84}} & \multicolumn{1}{c|}{\textbf{12.84}} & \textbf{2.30}  & \multicolumn{1}{c|}{\textbf{91.63}} & \multicolumn{1}{c|}{\textbf{90.60}} & \textbf{42.91} \\ \hline
\end{tabular}
\end{table}

% \usepackage{multirow}
\begin{table}[t!]
%\captionsetup{font=scriptsize}

\caption{Quantitative result of our method (PartA2) on Nuscenes OOD benchmark.}
\centering
\scriptsize
\label{NUSC_Part}
\begin{tabular}{|c|c|ccc|ccc|}
\hline
\multirow{2}{*}{Class}       & Method        & \multicolumn{3}{c|}{Recall @IOU}                                                           & \multicolumn{3}{c|}{OOD detection}                                                         \\ \cline{2-8} 
                                          & PartA2  & \multicolumn{1}{c|}{0.10}           & \multicolumn{1}{c|}{0.25}           & 0.40           & \multicolumn{1}{c|}{AUC$\uparrow$}            & \multicolumn{1}{c|}{AP$\uparrow$}             & FPR$\downarrow$            \\ \hline
\multirow{2}{*}{Deb}                   & Base          & \multicolumn{1}{c|}{36.94}          & \multicolumn{1}{c|}{15.91}          & 1.96           & \multicolumn{1}{c|}{70.06}          & \multicolumn{1}{c|}{59.75}          & 69.55          \\
                                          & \textbf{Ours} & \multicolumn{1}{c|}{\textbf{45.88}} & \multicolumn{1}{c|}{\textbf{33.14}} & \textbf{15.29} & \multicolumn{1}{c|}{\textbf{93.77}} & \multicolumn{1}{c|}{\textbf{95.46}} & \textbf{41.57} \\ \hline
\multirow{2}{*}{PPO} & Base          & \multicolumn{1}{c|}{24.32}          & \multicolumn{1}{c|}{13.32}          & 3.67           & \multicolumn{1}{c|}{60.66}          & \multicolumn{1}{c|}{52.54}          & 81.66          \\
                                          & \textbf{Ours} & \multicolumn{1}{c|}{\textbf{28.71}} & \multicolumn{1}{c|}{\textbf{19.08}} & \textbf{10.02} & \multicolumn{1}{c|}{\textbf{92.08}} & \multicolumn{1}{c|}{\textbf{94.13}} & \textbf{53.95} \\ \hline
\multirow{2}{*}{TB}          & Base          & \multicolumn{1}{c|}{29.21}          & \multicolumn{1}{c|}{3.65}           & 0.81           & \multicolumn{1}{c|}{69.18}          & \multicolumn{1}{c|}{58.29}          & 68.15          \\
                                          & \textbf{Ours} & \multicolumn{1}{c|}{\textbf{39.76}} & \multicolumn{1}{c|}{\textbf{19.07}} & \textbf{6.49}  & \multicolumn{1}{c|}{\textbf{94.57}} & \multicolumn{1}{c|}{\textbf{96.11}} & \textbf{36.51} \\ \hline
\multirow{2}{*}{TC}             & Base          & \multicolumn{1}{c|}{23.51}          & \multicolumn{1}{c|}{8.67}          & 1.73           & \multicolumn{1}{c|}{62.79}          & \multicolumn{1}{c|}{55.12}          & 87.09          \\
                                          & \textbf{Ours} & \multicolumn{1}{c|}{\textbf{30.38}} & \multicolumn{1}{c|}{\textbf{14.04}} & \textbf{5.58}  & \multicolumn{1}{c|}{\textbf{91.43}} & \multicolumn{1}{c|}{\textbf{93.58}} & \textbf{62.12} \\ \hline
\multirow{2}{*}{Ani}                   & Base          & \multicolumn{1}{c|}{20.31}          & \multicolumn{1}{c|}{1.53}           & 0.00           & \multicolumn{1}{c|}{70.60}          & \multicolumn{1}{c|}{59.72}          & 66.86          \\
                                          & \textbf{Ours} & \multicolumn{1}{c|}{\textbf{42.26}} & \multicolumn{1}{c|}{\textbf{17.02}} & \textbf{4.97}  & \multicolumn{1}{c|}{\textbf{94.53}} & \multicolumn{1}{c|}{\textbf{95.97}} & \textbf{34.61} \\ \hline
\end{tabular}
\end{table}

\section{Implementation Details}

\subsection{SUN-RGBD Dataset Details}
% 기존의 research를 따라 SUN-RGBD를 3D object detection training 에 활용하기 위해 형성된
% database는 다음과 같이 10가지 class [chair, desk, table, bookshelf, bed, night_stand, dresser, sofa, toilet, bathub] 로 구성되어있고 각각은  [9279, 933, 2539, 204, 771, 293, 182, 706, 171, 67] 개의 sample로 구성된다. 우리는 이를 class에 따라 5:5로 train 및 test split 하여 각각을 training 및 test (benchmark 구성) 을 위해 사용한다.
% 우리는 이중에  [chair, desk, table, bookshelf, bed] class를 anomaly sample augmentation을 위한 database로 활용한다. 이들의 sample 개수는 총 13725 개이다
% 우리는 이중에  [night_stand, dresser, sofa, toilet, bathub] 를 synthetic benchmark 구성을 위한 database 로 활용한다. 이들의 sample 개수는 총 1419 개 이다
\begin{itemize}

 \item \noindent The in-door SUN-RGBD database formed for 3D object detection in previous research, consists of 10 classes: [chair, desk, table, bookshelf, bed, nightstand, dresser, sofa, toilet, bathtub]. Each class is composed of the following number of samples: [9279, 933, 2539, 204, 771, 293, 182, 706, 171, 67]. 

  \item \noindent For anomaly sample augmentation, we specifically use the classes [chair, desk, table, bookshelf, bed], which collectively have a total of 13,725 samples. On the other hand, the classes [nightstand, dresser, sofa, toilet, bathtub], with a total of 1,419 samples, are utilized for constructing the SUN-RGBD OOD benchmark.
\end{itemize}

% Database chair: 9278
% Database desk: 933
% Database table: 2539
% Database bookshelf: 204
% Database bed: 771\\
% 1, 9279, 10211, 12751, 12955
% Database night_stand: 293
% Database dresser: 182
% Database sofa: 706
% Database toilet: 171
% Database bathtub: 67

\subsection{Configuration Setting Details}
% 우리는 먼저 training configuration 관련하여 본문에 다루지 못한 자세한 내용을 보충한다. 추가적으로 'Anomaly' class를 추가하였고
% 이와 관련하여 기존의 3가지 anchor box [[3.9, 1.6, 1.56]] [[0.8, 0.6, 1.73]] [[1.76, 0.6, 1.73]] 를 사용한 것에 비해 우리는 'Anomaly' class의  various size를 handle하기 위하여 추가적으로 anchor box size 9가지를 구성한다. 이러한 'Anomaly' class에 대해서 anchor rotation 과 anchor bottom height 그리고 matched threshold 및 unmatched threshold는 원래의 Pedestrian 의 configuration setting과 동일하게 하였다.
% 다음으로 우리는 원래의 OpenPcdet의 configuration을 따라 SECOND, PointPillars 그리고 PartA2에 대해서는batchsize per GPU를 4로하였고, PV-RCNN에 대해선 2로 설정하였다. PC 환경 관련하여   Quadro RTX 5000 16GB 8개를 갖는 GPU 환경에서 실험을 진행하였다.   

\begin{itemize}

\item \noindent Firstly, let's supplement additional details regarding the training configuration that are not covered in the main text. We introduce an additional 'Anomaly' class, and in connection with this, unlike the original three sizes of anchor boxes [[3.9, 1.6, 1.56]], [[0.8, 0.6, 1.73]], [[1.76, 0.6, 1.73]], we configure nine additional anchor box sizes to handle various sizes of the 'Anomaly' class. For this 'Anomaly' class, the anchor rotation, anchor bottom height, and matched/unmatched thresholds are set identically to the original configuration of the Pedestrian class.
Specifically, the additional 9 sizes of anchor boxes are as follows: [[1.76, 0.6, 0.87], [0.8, 0.6, 0.87], [3.9, 1.6, 0.78], [3.9, 1.6, 1.56], [5.15, 1.91, 2.19], [9.20, 2.61, 3.36], [15.56, 2.36, 3.53], [2.5, 1.24, 1.62], [1.06, 0.54, 1.28]].
                %'anchor_sizes': [[1.76, 0.6, 0.87],[0.8, 0.6, 0.87],[3.9, 1.6, 0.78],[3.9, 1.6, 1.56],[5.15, 1.91, 2.19],[9.20, 2.61, 3.36],[15.56, 2.36, 3.53],[2.5, 1.24, 1.62],[1.06, 0.54, 1.28]],

\item \noindent Next, following the original configuration of OpenPCDet, we set the batch size per GPU to 4 for SECOND, PointPillars, and PartA2, while for PV-RCNN, it is set to 2. The experiments are conducted in an environment equipped with eight Quadro RTX 5000 16GB GPUs.

% 다음으로, 우리의 dataset configuration 관련하여 본문에 다루지 못한 자세한 내용을 보충한다.
% 우리는 기존의 baseline detector에 대한 gt sample augmentation 에 더불어 추가적인 Anomaly sample augmentation을 구성한다. 구체적으로, 우리는 기존의 SUN-RGBD의 Anomaly sample에 KITTI database의 물체의 location 만을 치환하여 Anomaly database를 구성한다. 또한 이과정에서 Multi-size Mix augmentation을 수행하기 위해 그 중 짝수번째는 원래의 original 한 anomaly sample을 홀수번째에 대해서는 Resize 된 anomaly sample로 구성하였다.  
% 구성된 Anomaly database로 부터 filter by min pont 는 기존 gt sample과 같이 5로 설정하고, sample의 개수 또한 Car에 대한 gt sample과 같이 20으로 설정하고 나머지는 모두 기존의 gt sample setting을 따른다.

\item \noindent Furthermore, we present additional details regarding our dataset configuration, which are not covered in the main text. In addition to the ground-truth sample augmentation for the original baseline detector, we incorporate additional Anomaly sample augmentation. Specifically, we construct the Anomaly database by substituting only the location of KITTI database objects into the original SUN-RGBD Anomaly samples. Moreover, to perform Multi-size Mix augmentation in this process, we organize even-numbered samples as the original anomaly samples and odd-numbered samples as resized anomaly samples.

\item \noindent  From the constructed Anomaly database, we set the "filter by min point" to 5, consistent with the setting for the original ground-truth samples, and the copy-pasted "sample number" is set to 20 consistent with the original augmentation of Car ground-truth "sample number".
\end{itemize}

\subsection{Hyperparameter Setting}

% 본문의 공간제약으로 다루지못한 자세한 hyperparameter setting 에 대한 내용을 보충한다.
% 먼저 loss parameter 와 관련하여 $\lambda_{en}=1.0$ 그리고 $\lambda_{c}=1.0$ 으로 자세한 tuning없이 basic한 setting으로 설정하였다. energ loss 및 contrastive loss에 대해 각각 자세히 설명한다. Detection에서는 batch-size 에 따라 input으로 받는 scene 의 개수가 정해진다. 우리는 이렇게 각각의 scene 안에서 여러개의 ID object 와 'Anomaly' class object 들이 수집된다. 이렇게 수집된 ID object 와 OOD object에 대해 energy regularization loss 및 contrastive loss 계산을 수행한다.  
% energy loss 관련하여 E(x)의 계산에 있어서 'Anomaly' class 를 제외한 3개 class: Car, Ped, Cyclist 세개에 대한 logsumexp 를 활용한다. $m_{in}=-6.0$ $m_{out}=-3.0$ 으로 설정하였다. 
% contrastive loss 관련하여 $\tau_{c}=0.10$ 으로 기존의 contrastive learning setting에서 주로 쓰는 값을 사용하였다. 
% objectness loss 는 기존의 classification loss 와 동일한 weight로 추가적으로 objectness node 를 학습한다 
% 'Anomaly' class를 포함한 모든 foreground 에 대해 1 나머지에 대해 0을 주어 학습한다.

\begin{itemize}

\item \noindent To supplement the detailed hyperparameter settings not covered due to space constraints in the main text, we delve into further details.

\item \noindent Firstly, regarding loss parameters, we set $\lambda_{en}=1.0$ and $\lambda_{c}=1.0$ without fine-tuning for a basic configuration. We provide detailed explanations for both the energy loss and contrastive loss. In detection, the number of scenes processed as input is determined by the batch size. Within each scene, multiple ID objects and 'Anomaly' class objects are collected. For these collected ID (seen) objects and OOD (unseen) objects, we compute energy regularization loss and contrastive loss.

\item \noindent For energy loss calculation $E(x)$, we calculate log-sum-exp for three classes excluding the 'Anomaly' class: Car, Pedestrian, and Cyclist. We set $m_{in}=-6.0$ and $m_{out}=-3.0$.

\item \noindent Regarding contrastive loss, we use $\tau_{c}=0.10$, a commonly used value in traditional contrastive learning settings.

\item \noindent Objectness loss is added with the same weight as the existing classification loss, introducing additional learning for the objectness node. It is trained to assign 1 for all foreground, including the 'Anomaly' class, and 0 for the rest.
\end{itemize}
% 우리는 기존의 dense head network에 대해 추가적인 network를 구성한다
% 각각은 objectness node 에 대한 구성과 contrastive embedding을 위한 embedding layer이다. 
% 먼저, 기존 classification node는 1*1 convolution으로 다음과 같이 Conv2D [Input channels, anchorperlocation * numberofclass ] 구성된다.
% 우리의 objectness node는 anchor 별로 1개의 노드로 다음과같이 Conv2D [Input channels, anchorperlocation  ] 으로 구성된다.
% 우리의 contrastive embedding 은 anchor 별로 contrastive feature 를 얻는 방식으로
% Conv2D [Input channels, anchorperlocation * featuresize  ] 의 형태로 구성된다
% 여기서 우리는 featuresize 를 GPU 메모리를 고려하여 64로 설정하였다
% 여기서 얻어진 contrastive embedding을 L2 normalize 하여 contrastive learning에 활용한다.

\subsection{Additional Network Details}

\begin{itemize}
\item \noindent We introduce additional networks for the existing dense head network, each dedicated to the objectness node configuration and an embedding layer for contrastive embedding.

\item \noindent Firstly, the existing classification node is composed of a $1\times1$ convolution as follows:

Conv2D [Input-channels, anchor-per-location $\times$ number-of-classes].

\item \noindent Our objectness node is configured with one node per anchor as follows: Conv2D [Input-channels, anchor-per-location].

\item \noindent The contrastive embedding in our approach obtains contrastive features for each anchor. It is structured as 

Conv2D [Input-channels, anchor-per-location $\times$ feature-size]. Here, we set the feature size to 64, considering GPU memory.

\item \noindent  The obtained contrastive embedding is L2 normalized and utilized for contrastive learning.
\end{itemize}
\subsection{Synthetic Benchmark Construction Setting}

% Synthetic benchmark 는 실제 training 에서와 같이 copy-paste augmentation을 수행하지만 이렇게 합성된 scene 에 대해 test하는 방식이다.
% 실제 우리는 inference 과정에 대해 기존의 dataset configuration을 변형하여 사용한다. 
% 구체적인 방식은 기존의 anomaly sample augmentation과 동일하다. 다만, 우리는 새로이 SUN-RGBD의 test split 을 활용하여 별도의 synthetic database으로부터 sample을 얻는다. 구체적으로, 우리는 고정된 benchmark 구성을 위해
% 한번만 random하게 augmentation 을 통해 얻고 이를 저장하여 고정적으로 사용한다.
% 더나아가, 우리는 synthetic benchmark 와 misc bechmkark 와 OOD sample수를 비슷하게 맞추기 위해
% misc 가 있는 scene 에 대해서만 합성하는 synthetic sample 의 개수 1로 설정하였다. 이 때, 기존 validation set 에서 misc를 포함한 기존의 나머지 foreground class들은 모두 제거하고 syn sample을 합성하여 넣었다.
% synthetic database 구성 관련하여 우리는 KITTI gt database를 target object로 하여 NN grid sampling을 수행하며 hyperparameter인 slice number $N=5$ 로 설정한다. 또한, KITTI database의 물체의 location 만을 치환하여 구성한다.
% 여기서도 마찬가지로, Misc benchmark 와 동일하게 합성된 결과 중 0~50m 에 해당하는 UFO가 존재하는 scene만을 선택하여 validation 한다. 
\begin{itemize}
\item \noindent The synthetic benchmark employs copy-paste augmentation, similar to the actual training process, but tests on synthetically generated scenes. In practice, we modify the existing dataset configuration for the inference process. The specific approach aligns with the previous anomaly sample augmentation, where we use a test split from SUN-RGBD to construct a separate synthetic database from training. To ensure a consistent benchmark setup, we perform random augmentation only once and save it for fixed and repeated use. For Nuscenes OOD benchmark, we use the original training Nuscenes database for augmentation. 

\item \noindent To balance the number of unseen object samples between the synthetic benchmark and the KITTI Misc benchmark, we set the copy-pasting number of synthetic samples to 1 for scenes with miscellaneous objects. Specifically, for these scenes, we remove all other foreground classes, including miscellaneous objects, from the existing validation set and insert the synthetically generated samples.

\item \noindent Concerning the construction of the synthetic database, we use the KITTI ground truth database as the target object for NN grid sampling, with a slice number $N=5$. Similarly, we construct the synthetic database by substituting only the location of KITTI database objects into the original SUN-RGBD samples(or, Nuscenes samples).

\item \noindent Here as well, similar to the Misc benchmark, we select scenes with unseen objects within the range of 0-50m from the synthesized results for validation.
\end{itemize}

\section{Original 3D Object Detection Results}

% cifar 10 (q=0.75) / 10 LT (q=0.8) 는 one step
% cifar 100 (q=0.85) / 100 LT (q=0.9) 는 no step

\begin{table*}[t!]
%\captionsetup{font=scriptsize}

\caption{Detection 3D AP result of our method on the KITTI dataset.}
\centering
\scriptsize
\label{Supp_table_detection}

\begin{tabular}{|cc|ccc|ccc|lll|}
\hline
\multicolumn{2}{|c|}{\multirow{2}{*}{METHOD}}                       & \multicolumn{3}{c|}{Car 3D AP (R40)}                                        & \multicolumn{3}{c|}{Pedestrian 3D AP (R40)}                                        & \multicolumn{3}{l|}{Cyclist 3D AP (R40)}                                             \\ \cline{3-11} 
\multicolumn{2}{|c|}{}                                              & \multicolumn{1}{c|}{Easy}      & \multicolumn{1}{c|}{Mod}       & Hard      & \multicolumn{1}{c|}{Easy}      & \multicolumn{1}{c|}{Mod}       & Hard      & \multicolumn{1}{c|}{Easy} & \multicolumn{1}{c|}{Mod} & \multicolumn{1}{c|}{Hard} \\ \hline
\multicolumn{1}{|c|}{\multirow{2}{*}{SECOND}}       & Base          & \multicolumn{1}{c|}{91.26}          & \multicolumn{1}{c|}{81.52}          &    78.57       & \multicolumn{1}{c|}{57.88}          & \multicolumn{1}{c|}{51.84}          &  47.58         & \multicolumn{1}{l|}{83.14}     & \multicolumn{1}{l|}{66.29}    &         61.94                  \\
\multicolumn{1}{|c|}{}                              & \textbf{Ours} & \multicolumn{1}{c|}{91.32 } & \multicolumn{1}{c|}{81.51 } & 78.38  & \multicolumn{1}{c|}{58.74 } & \multicolumn{1}{c|}{53.33} & 49.47  & \multicolumn{1}{l|}{81.23}     & \multicolumn{1}{l|}{63.78}    &   60.74                       \\ \hline
\multicolumn{1}{|c|}{\multirow{2}{*}{PointPillars}} & Base          & \multicolumn{1}{c|}{88.76}          & \multicolumn{1}{c|}{78.71}          &   75.67        & \multicolumn{1}{c|}{55.53}          & \multicolumn{1}{c|}{49.56}          &    44.93       & \multicolumn{1}{l|}{80.03}     &  \multicolumn{1}{l|}{61.45}    &   57.19                        \\
\multicolumn{1}{|c|}{}                              & \textbf{Ours} & \multicolumn{1}{c|}{ 87.83} & \multicolumn{1}{c|}{ 75.39} & 72.15  & \multicolumn{1}{c|}{53.30 } & \multicolumn{1}{c|}{47.71 } & 43.05  & \multicolumn{1}{l|}{81.24}     & \multicolumn{1}{l|}{63.00}    &       58.76                    \\ \hline
\multicolumn{1}{|c|}{\multirow{2}{*}{PV-RCNN}}      & Base          & \multicolumn{1}{c|}{92.95}          & \multicolumn{1}{c|}{85.16}          &    82.41      & \multicolumn{1}{c|}{69.04}          & \multicolumn{1}{c|}{60.60}          &  54.97         & \multicolumn{1}{l|}{91.68}     & \multicolumn{1}{l|}{71.90}    &  67.31                      \\
\multicolumn{1}{|c|}{}                              & \textbf{Ours} & \multicolumn{1}{c|}{ 92.35} & \multicolumn{1}{c|}{84.50 } &  82.07   & \multicolumn{1}{c|}{60.27 } & \multicolumn{1}{c|}{54.48 } & 50.29  & \multicolumn{1}{l|}{85.26}     & \multicolumn{1}{l|}{66.68}    &    62.72                         \\ \hline
\multicolumn{1}{|c|}{\multirow{2}{*}{Part A2}}      & Base          & \multicolumn{1}{c|}{92.30}          & \multicolumn{1}{c|}{84.06}          &    81.68       & \multicolumn{1}{c|}{62.22}          & \multicolumn{1}{c|}{56.37}          &   51.19        & \multicolumn{1}{l|}{89.41}     & \multicolumn{1}{l|}{72.09}    &     67.47                      \\
\multicolumn{1}{|c|}{}                              & \textbf{Ours} & \multicolumn{1}{c|}{92.74} & \multicolumn{1}{c|}{84.43 } &  82.18 & \multicolumn{1}{c|}{66.10 } & \multicolumn{1}{c|}{57.68 } & 52.12  & \multicolumn{1}{l|}{89.83}     & \multicolumn{1}{l|}{70.71}    &    66.65                       \\ \hline
\end{tabular}

\end{table*}

%\input{table_supp_latex/UFO_detection_BEV_table}

% 우리는 기존 baseline 과 우리의 방법에 대해 원래의 3D object detection performance를 비교한다. Tabel에 나와있듯이, 우리의 방법은 기존의 baseline에 비해 detection AP에 있어서 minor 한 손실이 있다.
% 기존의 classification task에 대해서도 OOD detection performance의 향상에는 항상 classification accuracy가 떨어지는 현상과 일맥상통한다.
% 주요한 원인으로, 우리의 방법은 car, pedestrian 및 cyclist를 제외하고도 scene에 mix되어있는 Unidentified Foreground Object 에 대해서도 high confidence를 준다는 점이 있다. 기존의 detection AP에 대한 손실을 최대한 줄이기 위한 연구를 future work로 남겨둔다.

We compare the original 3D object detection performance between the baseline and our approach. As shown in Table~\ref{Supp_table_detection}, our method exhibits only minor loss in detection AP compared to the baseline. The observed decrease in detection AP is similar to the trade-off between classification accuracy and OOD classification performance in the classification task. A major contributing factor is that our method assigns high confidence to unidentified foreground objects mixed in the scene, beyond in-distribution classes: Car, Pedestrian, and Cyclist. We leave further research to minimize the loss in the original detection AP as future work.

\section{More Visualization of Result on KITTI Misc Benchmark}
\begin{figure*}[h!]
    \centering
    \includegraphics[width=\linewidth]{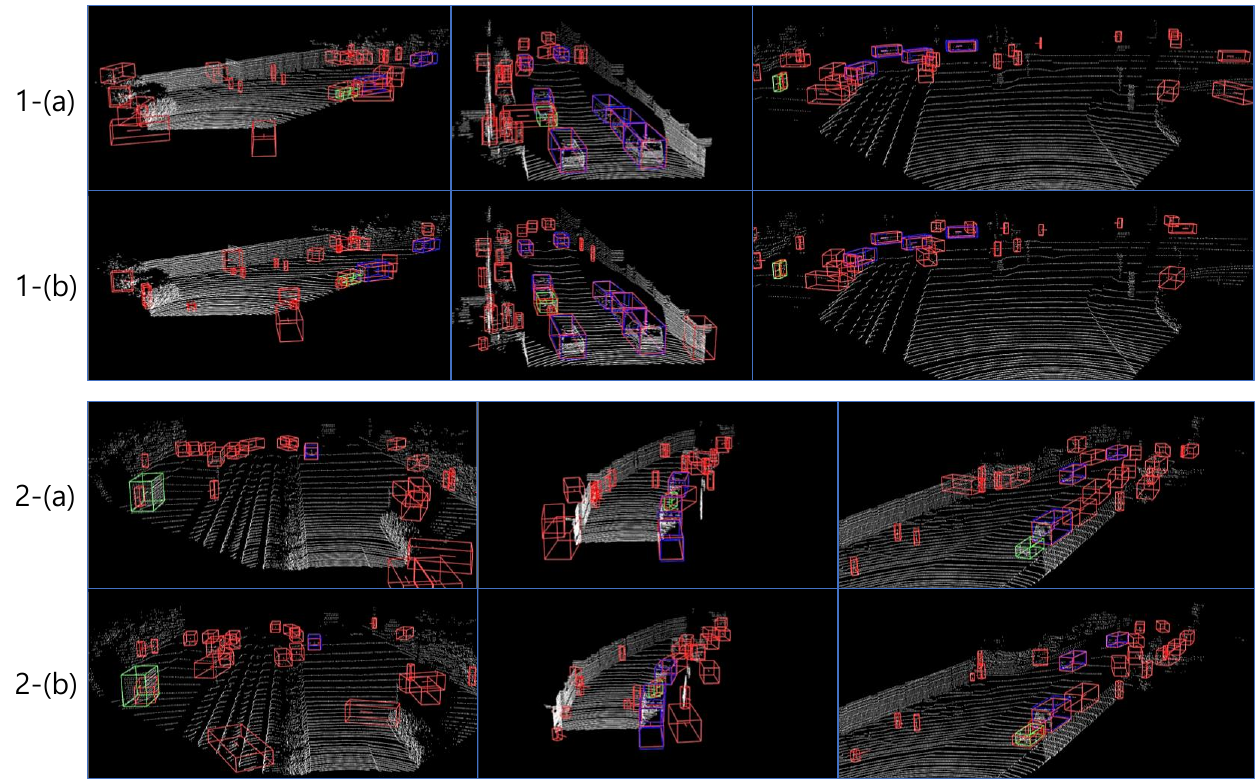}
    \caption{\textbf{More Qualitative result of our method on KITTI Misc benchmark.} (a): Base detector result; (b): Our result.}
    % \label{num_samples}
    \label{fig_supp_misc}
\end{figure*}
% 우리는 공간제약으로 다루지 못한 추가적인 visualization results for the KITTI Misc benchmark 를 다룬다.
% Figure 에서 보듯이 우리의 방법은 다양한 위치와 형태의 Misc object에 대해 기존의 방벙에 비해 더 정밀한 localization 을 보여준다. 우리의 방법의 유효성을 더 다양한 scene에 대해 검증하였다.

We address additional visualization results for the KITTI Misc benchmark that couldn't be covered due to space constraints. As shown in Figure~\ref{fig_supp_misc}, our approach demonstrates more precise localization for various positions and shapes of Misc objects compared to the baseline. We have further validated the effectiveness of our method across a diverse range of scenes.

\section{Visualization of Result on Synthetic Benchmark}
\begin{figure*}[h!]
    \centering
    \includegraphics[width=\linewidth]{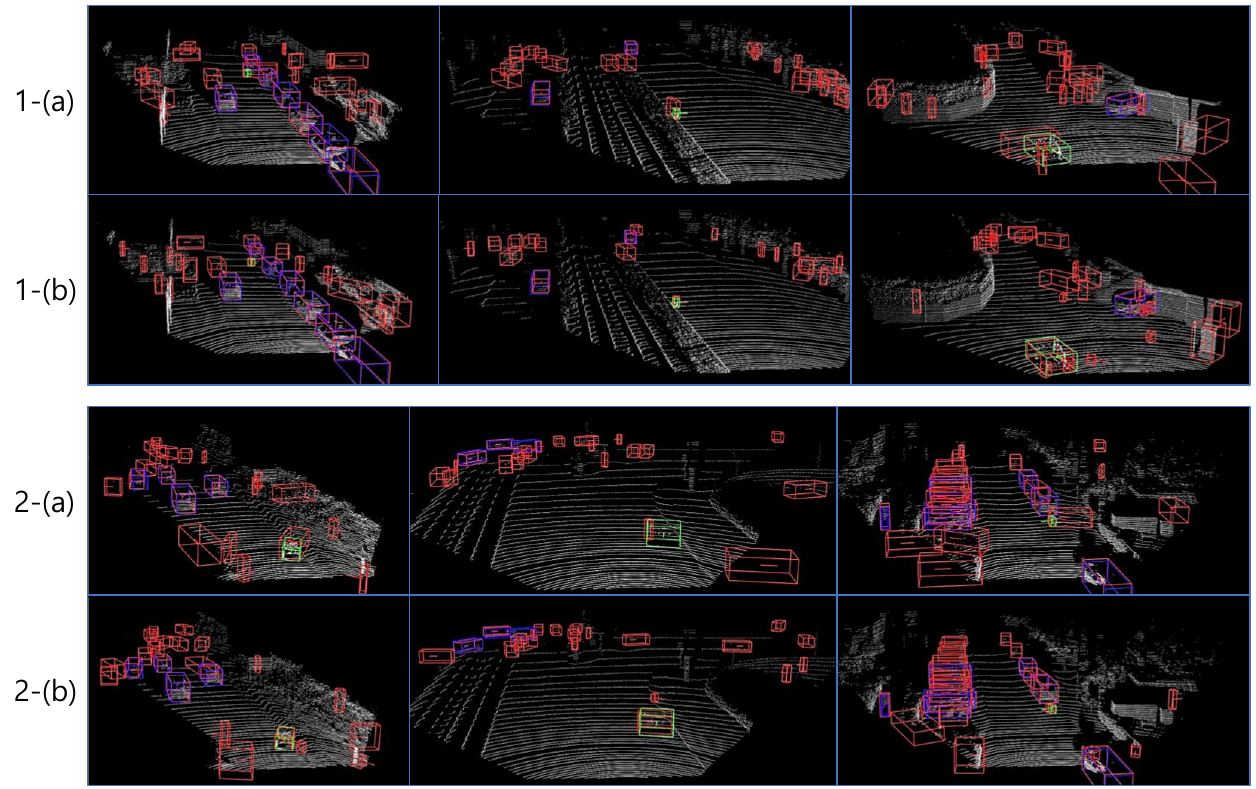}
    \caption{\textbf{Qualitative result of our method on sythetic benchmark.} (a): Base detector result; (b): Our result.}
    % \label{num_samples}
    \label{fig_supp_syn}
\end{figure*}

We present visualization results for the synthetic benchmark that couldn't be covered due to space constraints. The synthetic benchmark models a more diverse range of unidentified foreground objects in terms of size and shape compared to the traditional Misc benchmark. As depicted in Figure~\ref{fig_supp_syn}, our approach exhibits more precise localization for various positions and shapes of unidentified foreground objects compared to the baseline. We qualitatively validate the effectiveness of our method on the synthetic benchmark. Performing 3D detection and OOD classification for these unseen objects is an essential technology for ensuring the stability of real autonomous driving algorithms.

\section{Limitations with Visualization}
\begin{figure*}[h!]
    \centering
    \includegraphics[width=\linewidth]{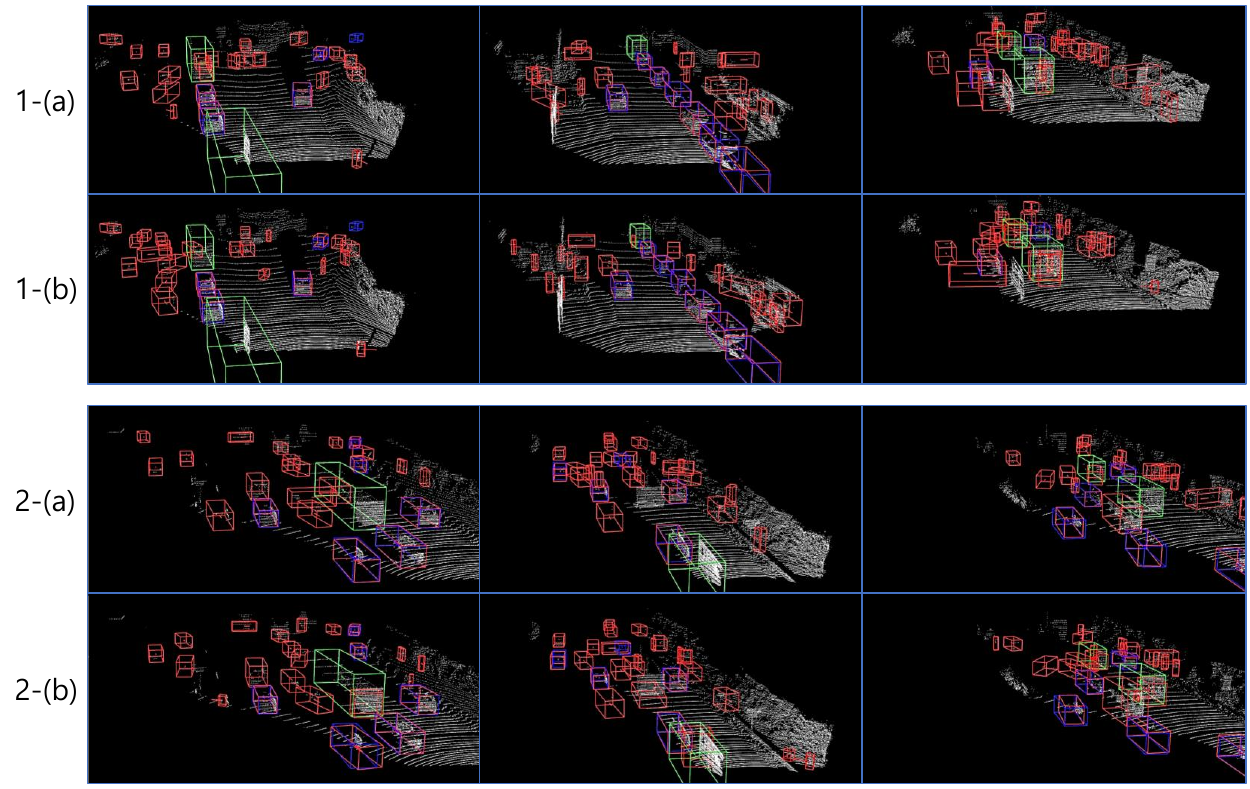}
    \caption{\textbf{Visualization result of failure case on KITTI Misc benchmark.} (a): Base detector result; (b): Our result.}
    % \label{num_samples}
    \label{fig_supp_fail}
\end{figure*}
% 우리는 공간제약으로 다루지 못한 우리의 방법의 limitations를 visualization을 통해 분석하고자한다. 구체적으로, 우리는 UFO detection 에 있어서 failure case 들을 예시로 한계점을 찾아낸다. 
% Figure에서 보듯이, base detector와 더불어 우리의 방법 또한 커다란 UFO 에 대해 localization 함에 어려움을 겪는다. 또한, occlusion으로 sparse하여 충분한 context를 못 갖는 UFO 에 대해서 역시 localization에 실패하고있다. 이는 기존의 3D object detector 들도 비슷하게 겪고 있는 문제로 큰 물체 또는 point가 sparse하여 context가 부족한 물체에 대해 detection 하는 문제는 future work로 남겨둔다. 

We aim to analyze the limitations of our method through visualization, which couldn't be covered due to space constraints. Specifically, we identify the shortcomings by showcasing failure cases in unseen object detection. As seen in Figure~\ref{fig_supp_fail}, similar to the base detector, our method also struggles with localizing large unseen objects. Additionally, there are challenges in accurately localizing unseen objects with occlusion. This is because such objects have sparse points and lack sufficient context. This is a common issue experienced by conventional 3D object detectors, and detecting large objects with sparse points and limited context remains a future research challenge.

% \section{Societal Impact}
% \subsection{Broader Impacts}
%OOD 감지는 이상 감지의 한 분야입니다.
% OOD detection is a branch of anomaly detection.
% \newpage
% \input{table_supp_latex/UFO_NUSC_SECOND_table}

% \input{table_supp_latex/UFO_NUSC_PP_table}

\end{document}